\documentclass{article}

\PassOptionsToPackage{table}{xcolor}
\usepackage{iclr2027_conference,times}
\usepackage{amsmath,amssymb,mathtools,amsthm}
\newtheorem{lemma}{Lemma}
\newtheorem{theorem}{Theorem}
\theoremstyle{definition}
\newtheorem{definition}{Definition}
\theoremstyle{plain}
\usepackage{algorithm}
\usepackage{multirow}
\usepackage{enumitem}
\usepackage{algorithmic}

\usepackage{booktabs}
\usepackage{tabularx}
\usepackage{siunitx}
\usepackage{nicematrix}
\usepackage{tikz}
\usepackage{graphicx}
\usepackage{xcolor}
\usepackage{placeins}
\usepackage{flafter}
\usepackage{hyperref}
\usepackage{url}

\definecolor{resultaccent}{HTML}{197BD1}
\definecolor{resulttext}{HTML}{174F86}
\definecolor{resultshade}{HTML}{F0F7FF}
\definecolor{resultheader}{HTML}{D8EBFC}
\definecolor{resultwinner}{HTML}{C7E3FB}
\definecolor{winnergray}{HTML}{E9EDF1}
\definecolor{tablerule}{HTML}{C8D0DA}
\definecolor{tablemuted}{HTML}{626B76}
\newcommand{\method}{\textsc{DWE}}
\title{Distillation with Errors:\\Sparse Unbiased Backpropagation\\for Knowledge Distillation}
\hypersetup{pdftitle={Distillation with Errors: Sparse Unbiased Backpropagation for Knowledge Distillation},pdfauthor={Yuyang Zhang and Yuxin Li},pdfkeywords={Knowledge Distillation, Sparse Backpropagation, Large Language Models, Unbiased Gradient Estimation}}
\author{\parbox[t]{\dimexpr\textwidth-2\tabcolsep\relax}{\centering
\textbf{Yuyang Zhang\textsuperscript{*,\textdagger}\qquad Yuxin Li\textsuperscript{*}}\\[3pt]
\normalfont University of Science and Technology of China\\[2pt]
\texttt{\{yuyzhang027,yuxinli118\}@gmail.com}
}}
\iclrfinalcopy
\usepackage{booktabs}
\usepackage{siunitx}
\usepackage{nicematrix}
\usepackage{tikz}
\usepackage{graphicx}
\usepackage{enumitem}
\usepackage{wrapfig}

\definecolor{ink}{HTML}{15181F}
\definecolor{gray800}{HTML}{31363D}
\definecolor{gray700}{HTML}{4C5058}
\definecolor{gray600}{HTML}{656971}
\definecolor{gray500}{HTML}{82868E}
\definecolor{gray400}{HTML}{A4A8AE}
\definecolor{gray300}{HTML}{C1C4C9}
\definecolor{gray200}{HTML}{D6D9DE}
\definecolor{gray150}{HTML}{E4E6EA}
\definecolor{gray100}{HTML}{EEEFF2}
\definecolor{gray50}{HTML}{F6F7F9}
\definecolor{blue800}{HTML}{133372}
\definecolor{blue700}{HTML}{1E4592}
\definecolor{blue600}{HTML}{2F59AB}
\definecolor{blue500}{HTML}{446FC0}
\definecolor{blue400}{HTML}{6F92D3}
\definecolor{blue300}{HTML}{99B3E3}
\definecolor{blue200}{HTML}{BFD0EF}
\definecolor{blue100}{HTML}{DCE6F7}
\definecolor{blue50}{HTML}{EDF2FB}
\definecolor{ochre700}{HTML}{966C1E}
\definecolor{ochre600}{HTML}{B98421}
\definecolor{ochre500}{HTML}{D69C32}
\definecolor{ochre400}{HTML}{E6B463}
\definecolor{ochre300}{HTML}{F0CB91}
\definecolor{ochre200}{HTML}{F6DEB9}
\definecolor{ochre100}{HTML}{FAEDD9}
\colorlet{dwemain}{blue600}
\colorlet{dwesoft}{blue300}
\colorlet{dwebase}{ochre600}
\colorlet{dwetint}{blue600!9}
\colorlet{dwemuted}{gray700}

\newlength{\dwecapskip}
\newcommand{\dwetable}[1][\small]{%
  \vspace{\dwecapskip}\centering#1%
  \renewcommand{\arraystretch}{1.22}%
  \setlength{\tabcolsep}{5.2pt}%
  \setlength{\aboverulesep}{0pt}\setlength{\belowrulesep}{0pt}%
  \setlength{\heavyrulewidth}{0.075em}\setlength{\lightrulewidth}{0.045em}\setlength{\cmidrulewidth}{0.035em}%
  \arrayrulecolor{ink}}

\newcommand{\dweours}{\rowcolor{dwetint}}

\newcommand{\dwemuted}[1]{{\color{dwemuted}#1}}

\newcommand{\dwehs}{\rule[-0.85ex]{0pt}{2.95ex}}
\newcommand{\dwenote}[1]{\par\vspace{3pt}{\footnotesize\dwemuted{#1}}}

\newcommand{\dwebar}[2][dwemuted]{%
  \makebox[2.9em][l]{\textcolor{#1}{\rule[0.08ex]{\dimexpr 2.9em*#2/1000\relax}{1.05ex}}}}

\colorlet{dweblue}{dwemain}
\colorlet{resulttext}{dwemain}
\colorlet{rklcoral}{gray500}
\colorlet{dweoutline}{blue300}
\colorlet{dwebest}{blue200}
\colorlet{inkmuted}{dwemuted}
\colorlet{rulegray}{gray400}

\colorlet{dwelight}{dwesoft}
\colorlet{dwerowtint}{dwetint}

\newcommand{\grouplabel}[1]{{\footnotesize\itshape\color{inkmuted}#1}}


\newsavebox{\dwetablebox}

\newcommand{\KDHeadSpeedMin}{1.63}
\newcommand{\KDHeadSpeedMax}{1.95}
\newcommand{\KDBudgetTimeFirst}{12.6}
\newcommand{\KDBudgetTimeLast}{15.5}
\newcommand{\KDBudgetTimeFull}{22.8}
\newcommand{\KDPairHiddenReduction}{33.6}
\newcommand{\KDPairHeadReduction}{23.0}
\newcommand{\KDGeometryHiddenReduction}{3.9}
\newcommand{\KDStratificationHiddenReduction}{42.8}
\newcommand{\KDStratificationHeadReduction}{44.9}
\newcommand{\KDOverallHiddenReduction}{63.5}
\newcommand{\KDOverallHeadReduction}{57.9}
\newcommand{\KDTrainingReduction}{15.7}
\newcommand{\KDFullTrainingHours}{1.54}
\newcommand{\KDDweTrainingHours}{1.30}
\newcommand{\KDScaleTrainingReduction}{10.4}
\newcommand{\OPDHeadSpeedMin}{1.96}
\newcommand{\OPDHeadSpeedMax}{2.20}
\newcommand{\OPDHeadExactEightK}{214.6}
\newcommand{\OPDHeadDweEightK}{97.7}
\newcommand{\OPDHeadBwdSavedEightK}{133.7}
\newcommand{\OPDHeadFwdAddEightK}{16.8}
\newcommand{\OPDAdjointBytesPerToken}{192}
\newcommand{\OPDAdjointMiBEightK}{1.5}
\newcommand{\OPDStepSavedPct}{1.53}
\newcommand{\OPDOverallExact}{38.95}
\newcommand{\OPDOverallIS}{39.18}
\newcommand{\OPDOverallDwe}{39.30}
\newcommand{\OPDMainAvgLine}{19}
\newcommand{\OPDMainLastLine}{20}
\begin{document}
\maketitle
\fancyhead{}
\ificlrfinal
\begingroup
\renewcommand{\thefootnote}{\fnsymbol{footnote}}
\footnotetext[1]{Equal contribution.\qquad\textsuperscript{\textdagger}Corresponding author.}
\endgroup
\fi
\begin{abstract}
Knowledge distillation (KD) trains a student to match a teacher's output
distribution. However, full-vocabulary KD requires costly output-head
backpropagation. To reduce this cost, we introduce \textbf{Distillation with
Errors (DWE)}. Our key insight is that minibatch training already
approximates the training gradient, so per-position backpropagation does not
need to be exact. DWE therefore exploits the zero-sum structure of logit
gradients to construct sparse paired gradient estimates whose expectation matches the
full-vocabulary gradient. It then uses geometry-aware pairing and stratified
sampling to further reduce estimation variance. We prove that the estimator
is unbiased and evaluate DWE's gradient estimation accuracy and training
efficiency. At the same sampling budget, DWE reduces hidden-state and
output-weight gradient MSE by \KDOverallHiddenReduction\% and
\KDOverallHeadReduction\%, respectively, compared with importance sampling.
In off-policy KD, it achieves up to a $\KDHeadSpeedMax\times$
output-head speedup and reduces training time by
\KDScaleTrainingReduction--\KDTrainingReduction\% relative to full-vocabulary
KD. In on-policy distillation (OPD), it achieves up to a $2.20\times$
output-head speedup.
DWE maintains competitive downstream performance in both settings.
\ificlrfinal
Our code is available at \url{https://github.com/dwe-lab/DWE}.
\fi
\end{abstract}
\begingroup
\setlength{\parskip}{4pt}
\section{Introduction}
\label{sec:introduction}

Knowledge distillation (KD)~\citep{hinton2015distilling,sanh2019distilbert,
gu2024minillm} transfers capabilities from a stronger teacher model to a
student by training the student to match the teacher's output distribution.
For modern language models~\citep{yang2025qwen3,grattafiori2024llama,
gemma2025gemma3} with large vocabularies, full-vocabulary KD computes a
dense logit gradient at each token position. Propagating this gradient
through the entire output head incurs substantial computational cost.

Existing approaches for reducing this backward computation broadly fall
into two categories. Top-$k$ gradient
selection~\citep{sun2017meprop} retains only the coordinates with the
largest gradient magnitudes. This reduces computation but introduces bias
by discarding the contributions of the remaining coordinates. In contrast,
importance sampling (IS)~\citep{wang2018atomo,oktay2021randomized} randomly selects coordinates
and reweights their contributions so that the estimated gradient matches
the full-vocabulary gradient in expectation. However, independent sampling
can leave substantial estimation error under a small sampling budget.
Can we reduce this backward computation while preserving the
full-vocabulary gradient in expectation and keeping the estimation error low?

To address this question, we propose \textbf{Distillation with Errors (DWE)},
a method for efficient output-head backpropagation in knowledge distillation.
Our key insight is that minibatch training already approximates the training
gradient, so per-position backpropagation does not need to be exact.
We therefore replace exact per-position backward
computation with a cheaper estimate that preserves the full-vocabulary
gradient in expectation.

DWE constructs this estimate in two stages. First, it uses the zero-sum
property of the logit gradient to form two sampling distributions from its
positive and negative parts. DWE samples one coordinate from each
distribution and gives them equal and opposite weights. This produces a
sparse, unbiased gradient estimate with lower variance than IS. Second, DWE uses
geometry-aware pairing and stratified sampling~\citep{douc2005comparison}
to further reduce variance.
Geometry-aware pairing couples coordinates with similar output rows,
reducing the error propagated to the hidden state. Stratified sampling
spreads $K$ draws across the sampling range to reduce variance.
Averaging the $K$ paired estimates limits output-head backpropagation to at
most $2K$ rows per token.
We prove that the resulting estimator is unbiased
(Appendix~\ref{app:stratified-unbiased}).

To assess its efficiency and generality, we evaluate DWE in both off-policy
(Section~\ref{sec:experiments}) and on-policy
(Appendix~\ref{app:opd-experiments}) distillation settings. At the same
sampling budget, DWE reduces hidden-state and output-weight gradient MSE
by \KDOverallHiddenReduction\% and \KDOverallHeadReduction\%, respectively,
compared with IS. DWE with 16 coordinate draws per token closely matches
the hidden-state gradient accuracy of IS with 44 draws.
In off-policy KD, DWE achieves a
$\KDHeadSpeedMin$--$\KDHeadSpeedMax\times$ speedup in total output-head
computation and reduces training time by
\KDScaleTrainingReduction--\KDTrainingReduction\% relative to
full-vocabulary KD, while maintaining comparable downstream performance.
In on-policy distillation (OPD)~\citep{agarwal2024gkd}, DWE achieves a
$1.96$--$2.20\times$ output-head speedup and reduces mean training-update
time by \OPDStepSavedPct\%, with competitive performance across 13 reasoning benchmarks.

Our main contributions are as follows.
\begin{enumerate}
\item We propose DWE, a method for efficient output-head backpropagation
in knowledge distillation. It produces sparse, unbiased gradient estimates
with lower estimation error than IS at the same sampling budget.

\item We thoroughly evaluate DWE in both off-policy KD and OPD.
At matched sampling budgets, DWE achieves lower gradient estimation error
than IS. Compared with full-vocabulary distillation, it reduces training
cost while maintaining comparable downstream performance.
\end{enumerate}
\section{Preliminaries}
\label{sec:preliminaries}

\begingroup
\setlength{\abovedisplayshortskip}{\abovedisplayskip}
\setlength{\belowdisplayshortskip}{\belowdisplayskip}

\paragraph{Knowledge distillation.}
KD~\citep{hinton2015distilling} trains a student model $\pi_\theta$ to
match a fixed teacher $\pi_{\mathrm T}$. In off-policy KD, training text
comes from a fixed dataset. For a token sequence
$y=(y_1,\ldots,y_T)$, let $s_t=y_{<t}$ denote the context at position $t$.
The student and teacher assign next-token probabilities
$p_t(v)=\pi_\theta(v\mid s_t)$ and
$q_t(v)=\pi_{\mathrm T}(v\mid s_t)$ over a shared vocabulary
$\mathcal V$ of size $V$.

We use the forward KL divergence as the token-level loss. For a sequence
$y$, the training objective averages these losses:
\begin{equation}
    \mathcal L(\theta;y)=\frac1T\sum_{t=1}^T\ell_t,
    \qquad
    \ell_t=D_{\mathrm{KL}}\!\left(q_t\,\Vert\,p_t\right)
      =\sum_{v\in\mathcal V}q_t(v)\log\frac{q_t(v)}{p_t(v)}.
    \label{eq:kd}
\end{equation}

In on-policy distillation (OPD)~\citep{agarwal2024gkd}, the student
generates a response $y\sim\pi_\theta(\cdot\mid x)$ to a prompt $x$,
giving contexts $s_t=(x,y_{<t})$. Our OPD experiments use reverse
KL~\citep{gu2024minillm,li2026rethinking,huang2026taopd}, whose logit
gradient is given in Appendix~\ref{app:kl-gradients}.

\begin{samepage}
\paragraph{Output-head gradients.}
At a token context $s$, the student maps its hidden state $h\in\mathbb R^d$
to logits $z=Wh$ using the output-head weights
$W\in\mathbb R^{V\times d}$. The student distribution is
$p=\operatorname{softmax}(z)$, and the teacher distribution is
$q=\pi_{\mathrm T}(\cdot\mid s)$.
For the token-level loss $\ell=D_{\mathrm{KL}}(q\|p)$, the logit gradient
$g=\nabla_z\ell$ has components
\begin{equation}
    g_v=p(v)-q(v).
    \label{eq:forward-kl-gradient}
\end{equation}
\end{samepage}

\noindent By the chain rule, the output-head backward pass computes the
hidden-state gradient and the output-weight gradient:
\begin{equation}
    \nabla_h\ell=W^\top g,\qquad
    \nabla_W\ell=g h^\top.
    \label{eq:head-backward}
\end{equation}
For $N$ token positions, these dense products require $O(NVd)$
arithmetic. DWE reduces this cost by replacing the dense logit gradients with
sparse unbiased estimates.
\endgroup
\par
\endgroup
\section{Distillation with Errors}
\label{sec:method}

\method{} reduces output-head backward computation in knowledge distillation
by replacing dense logit gradients with sparse estimates. Since training
already relies on minibatch gradient estimates, we need not backpropagate
the exact full-vocabulary gradient at every token position. \method{}
therefore constructs a sparse, unbiased estimate for each token position
and backpropagates only through the selected output rows.

\begin{figure}[!ht]
    \centering
    \includegraphics[width=\linewidth]{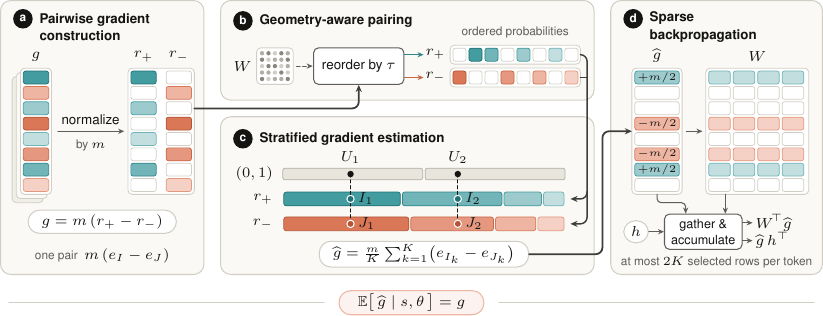}
    \setlength{\abovecaptionskip}{4pt}
    \caption{\textbf{Overview of \method{}.} The positive and negative parts of the
    logit gradient are normalized into two sampling distributions (a).
    Both distributions are reordered using the fixed geometric order
    $\tau$ (b), then sampled at shared, stratified quantiles to obtain
    $K$ coordinate pairs (c). Averaging the corresponding paired estimates
    gives a sparse gradient estimate that preserves the full gradient
    in expectation. This estimate enables backpropagation through only
    the selected output rows, at most $2K$ per token (d).
    The illustration uses $K=2$.}
    \label{fig:dwe-mechanism}
\end{figure}

\method{} constructs this estimate in three steps
(Figure~\ref{fig:dwe-mechanism}). First, \method{} uses the zero-sum
property of the logit gradient to construct two sampling distributions
from its positive and negative parts. It then samples from these
distributions to form sparse, unbiased paired gradient estimates.
Second, \method{} uses a fixed geometric ordering of the output rows to guide
the pairing and reduce hidden-state gradient error. Third, it uses
stratified sampling to distribute $K$ quantile draws across the sampling range and
reduce variance. With $K$ pairs, the sampling budget is $2K$ coordinate
draws per token, selecting at most $2K$ distinct output rows.

\subsection{Pairwise Gradient Construction}
\label{sec:sparse-gradient}

Pairwise gradient construction turns each dense logit gradient into a
two-sparse gradient estimate while preserving the original gradient in
expectation. The construction relies on the zero-sum property of softmax
gradients.

\begin{lemma}[Zero-sum logit gradients]
\label{lem:zero-sum}
Let $p=\operatorname{softmax}(z)$ and let $\ell(q,p)$ be differentiable
in $p$, with $q$ fixed. Let
$g=(g_1,\ldots,g_V)^\top=\nabla_z\ell(q,p)\in\mathbb R^V$
denote the logit gradient. Then
\[
    \sum_{v=1}^{V}g_v=0.
\]
\end{lemma}
The proof is given in Appendix~\ref{app:zero-sum}.

Write $(g_+)_v=\max(g_v,0)$ and $(g_-)_v=\max(-g_v,0)$, so that
$g=g_+-g_-$. By Lemma~\ref{lem:zero-sum}, the two parts have equal total
magnitude. This common magnitude is
\begin{equation}
    m=\sum_v(g_+)_v=\sum_v(g_-)_v=\frac12\|g\|_1.
    \label{eq:gradient-mass}
\end{equation}
For $g\neq0$, normalizing the two parts yields two probability distributions:
\begin{equation}
    r_+=\frac{g_+}{m},
    \qquad
    r_-=\frac{g_-}{m}.
    \label{eq:gradient-decomposition}
\end{equation}
These distributions are supported on the positive and negative coordinates,
respectively, and satisfy
\[
    g=m(r_+-r_-).
\]
For $g=0$, we use the zero estimate.

Sampling one coordinate from each distribution gives an unbiased two-sparse
estimate. Let $I\sim r_+$ and $J\sim r_-$. We assign $+m$ to $I$ and $-m$
to $J$, giving
\begin{equation}
    \widehat g_{\mathrm{pair}}=m(e_I-e_J),
    \label{eq:paired-update}
\end{equation}
where $e_i$ is the $i$-th standard basis vector. For any vocabulary
coordinate $v$,
\begin{equation}
    \begin{aligned}
    \mathbb E\!\left[(\widehat g_{\mathrm{pair}})_v\right]
        &=m\Pr(I=v)-m\Pr(J=v)\\
        &=m\bigl((r_+)_v-(r_-)_v\bigr)\\
        &=(g_+)_v-(g_-)_v=g_v.
    \end{aligned}
    \label{eq:sparse-representation}
\end{equation}
Each pair preserves equal positive and negative mass. In comparison,
IS samples coordinates independently with replacement using probabilities
$\rho_v=|g_v|/\|g\|_1$ and averages the reweighted contributions
$(g_v/\rho_v)e_v$. At the same sampling budget, pairing removes the variance
component caused by random sign allocation in IS
(Appendix~\ref{app:stratified-variance}). We next refine how coordinate
pairs are sampled to further reduce estimation error.

\subsection{Geometry-Aware Pairing}
\label{sec:geometry-pairing}

To reduce the error propagated to the hidden state, \method{} uses output-head
geometry to guide the sampling of coordinate pairs. We describe this joint
sampling by a coupling.

\begin{definition}[Coupling]
\label{def:coupling}
A coupling $\gamma$ of $r_+$ and $r_-$ is a joint distribution of $(I,J)$
whose marginals satisfy $I\sim r_+$ and $J\sim r_-$.
\end{definition}

All couplings with these marginals preserve the expected logit gradient in
Eq.~\eqref{eq:sparse-representation}. For a single paired estimate, they also
give the same mean-squared logit gradient error, but can yield different
hidden-state gradient errors after multiplication by $W^\top$.

Let $w_i^\top$ denote row $i$ of the output-head matrix $W$. A sampled pair
$(I,J)\sim\gamma$ produces the hidden-state gradient
$W^\top m(e_I-e_J)=m(w_I-w_J)$, whose expectation is the exact gradient
$W^\top g$. Its mean-squared error is therefore
\begin{equation}
    \mathbb E_{\gamma}\!\left[
        \left\|m(w_I-w_J)-W^\top g\right\|_2^2
    \right]
    =m^2\mathbb E_{\gamma}\!\left[\|w_I-w_J\|_2^2\right]
    -\|W^\top g\|_2^2.
    \label{eq:geometry-error}
\end{equation}
For fixed $g$ and $W$, the second term is independent of the coupling.
Thus, reducing the expected squared distance between paired output rows
reduces the mean-squared hidden-state gradient error.

Minimizing this distance over all couplings is an optimal transport problem.
Solving it for every token would be costly, so \method{} approximates the geometry
with a cached one-dimensional ordering of the vocabulary.
At initialization, we center the student
output-head rows and estimate their leading principal direction. We project
each row onto this direction and sort the vocabulary by its projection
score. The resulting order $\tau$ provides a one-dimensional proxy for the
output-head geometry, with $\tau(a)$ denoting the vocabulary coordinate at
position $a$.

\method{} then pairs positive and negative coordinates at the same
cumulative-probability level along this order. Let $F_{r_+,\tau}$ and
$F_{r_-,\tau}$ denote the cumulative distributions of $r_+$ and $r_-$ along
$\tau$. Using the same quantile $u\in(0,1)$ for both gives
\begin{equation}
    I(u)=\tau\!\left(F_{r_+,\tau}^{-1}(u)\right),
    \qquad
    J(u)=\tau\!\left(F_{r_-,\tau}^{-1}(u)\right).
    \label{eq:geometry-coupling}
\end{equation}
For the fixed one-dimensional projection scores, this shared-quantile
coupling minimizes the expected squared distance between paired
scores~\citep{peyre2019computational}. It also preserves the
marginals $I(U)\sim r_+$ and $J(U)\sim r_-$ for
$U\sim\operatorname{Unif}(0,1)$. The resulting paired estimate therefore
remains unbiased.

\subsection{Stratified Gradient Estimation}
\label{sec:stratified-estimation}

\method{} averages $K$ paired estimates to obtain a more stable gradient estimate.
It uses stratified sampling to spread these draws across the quantile
interval and further reduce variance.

With the shared-quantile pairing, the dense gradient can be expressed as an
integral over the quantile interval. Define the paired estimate
$f(u)=m(e_{I(u)}-e_{J(u)})$. Since $I(U)\sim r_+$ and $J(U)\sim r_-$ for
$U\sim\operatorname{Unif}(0,1)$,
\begin{equation}
    g=\mathbb E[f(U)]=\int_0^1 f(u)\,\mathrm du.
    \label{eq:coupling-integral}
\end{equation}

Independent draws from the full interval can concentrate in some regions
while leaving others unsampled. \method{} therefore divides $(0,1)$ into $K$
equal strata and draws one quantile independently from each:
\begin{equation}
    U_k\sim
    \operatorname{Unif}\!\left(
        \frac{k-1}{K},\frac{k}{K}
    \right),
    \qquad k=1,\ldots,K.
    \label{eq:stratified-sampling}
\end{equation}
Each quantile determines a pair $(I_k,J_k)=(I(U_k),J(U_k))$. Averaging the
corresponding estimates gives the \method{} estimator
\begin{equation}
    \widehat g=\frac{m}{K}\sum_{k=1}^{K}(e_{I_k}-e_{J_k}).
    \label{eq:dwe-estimator}
\end{equation}

Averaging the $K$ paired estimates gives an unbiased estimate of the integral
in Eq.~\eqref{eq:coupling-integral}, so \mbox{$\mathbb E[\widehat g]=g$}.
For a fixed coupling and $K$ pairs, stratified sampling has no greater
mean-squared error than independent quantile sampling.
Since each paired sample satisfies $\|f(u)\|_2^2=2m^2$, this gives
\begin{equation}
    \mathbb E\!\left[\|\widehat g-g\|_2^2\right]
    \leq \frac{2m^2-\|g\|_2^2}{K}.
    \label{eq:estimation-error-bound}
\end{equation}
The mean-squared error bound decreases as $1/K$, while the estimate has
at most $2K$ nonzero coordinates per token.
Proofs are given in Appendices~\ref{app:stratified-unbiased}
and~\ref{app:stratified-variance}.

\subsection{Sparse Backpropagation}
\label{sec:efficient-backward}

\method{} uses the sparse estimate constructed above to reduce output-head
backward computation while preserving the expected parameter gradient.
Since $\widehat g$ contains at most $2K$ nonzero coordinates, both output-head
backward products involve only the corresponding rows. Substituting
$\widehat g$ for the dense logit gradient gives
\begin{equation}
    W^\top\widehat g
    =\frac{m}{K}\sum_{k=1}^K(w_{I_k}-w_{J_k}),
    \qquad
    \widehat g h^\top
    =\frac{m}{K}\sum_{k=1}^K(e_{I_k}-e_{J_k})h^\top.
    \label{eq:sparse-backward}
\end{equation}
The first expression computes the hidden-state gradient by gathering only
the selected rows of $W$. The second accumulates the output-weight
gradient in the selected rows.

The resulting parameter gradient is also unbiased.
For a fixed context and parameter state, let
$\mathcal J_\theta=\partial z/\partial\theta$ denote the logit Jacobian.
Backpropagation with $\widehat g$ gives the parameter gradient
$\mathcal J_\theta^\top\widehat g$. By linearity,
\begin{equation}
    \mathbb E[\mathcal J_\theta^\top\widehat g\mid s,\theta]
    =\mathcal J_\theta^\top\mathbb E[\widehat g\mid s,\theta]
    =\mathcal J_\theta^\top g
    =\nabla_\theta\ell.
    \label{eq:parameter-unbiased}
\end{equation}
\subsection{Complexity Analysis}
\label{sec:complexity}

For $N$ supervised token positions, the two dense output-head
backward products operate over all $V$ rows, each of dimension $d$.
\method{} uses at most $2K$ rows per position. The dense and sparse
multiply-add counts are therefore
\begin{equation}
    C_{\mathrm{dense}}=2NVd,
    \qquad
    C_{\mathrm{DWE}}\leq4NKd.
    \label{eq:backward-cost}
\end{equation}
The arithmetic cost of these products decreases from $O(NVd)$ to
$O(NKd)$, with a reduction factor of at least
\begin{equation}
    \frac{2NVd}{4NKd}=\frac{V}{2K}.
    \label{eq:backward-reduction}
\end{equation}
\section{Experiments}
\label{sec:experiments}

We focus on off-policy KD to evaluate \method{}'s computational efficiency and
distillation quality. Additional results for OPD are reported in
Appendix~\ref{app:opd-experiments}.

\subsection{Experimental Setup}
\label{sec:experimental-setup}

\paragraph{Models and data.}
We evaluate distillation from Qwen3-4B-Instruct-2507~\citep{yang2025qwen3} to
Qwen3-0.6B~\citep{yang2025qwen3}, from Qwen3-8B~\citep{yang2025qwen3} to
Qwen3-1.7B~\citep{yang2025qwen3} and from Qwen3-32B~\citep{yang2025qwen3} to
Qwen3-4B~\citep{yang2025qwen3}. We train each
configuration on 80M tokens from
FineWeb-Edu~\citep{penedo2024fineweb}, packed into sequences of 4,096 tokens.
Within each configuration, all methods use the same cached teacher hidden
states. Training time excludes cache construction.

\paragraph{Methods.}
We compare \method{} with full-vocabulary KD (Full) and IS.
Full computes the exact logit gradient. IS samples coordinates
with replacement in proportion to their absolute gradient values and
reweights their contributions. We use $K=8$ pairs per token for \method{}
and $2K=16$ coordinate draws for IS. Within each configuration, all methods
optimize forward KL and start from the same student checkpoint, using the
same data order and optimization settings. Training uses a single NVIDIA
H100 80GB GPU.

\paragraph{Evaluation.}
We evaluate the final checkpoints on
HellaSwag~\citep{zellers2019hellaswag}, PIQA~\citep{bisk2020piqa},
ARC-Easy and ARC-Challenge~\citep{clark2018arc},
GSM8K~\citep{cobbe2021verifiers}, LAMBADA~\citep{paperno2016lambada},
and MMLU~\citep{hendrycks2021mmlu}. All methods use the complete evaluation sets and identical
zero-shot settings. We also measure forward KL and perplexity on held-out
text. Detailed training and evaluation settings for off-policy KD and OPD are
given in Appendix~\ref{app:experimental-configuration}.

\subsection{Output-Head Efficiency}
\label{sec:kd-head-efficiency}

\begin{figure}[!ht]
  \centering
  \includegraphics[width=\linewidth]{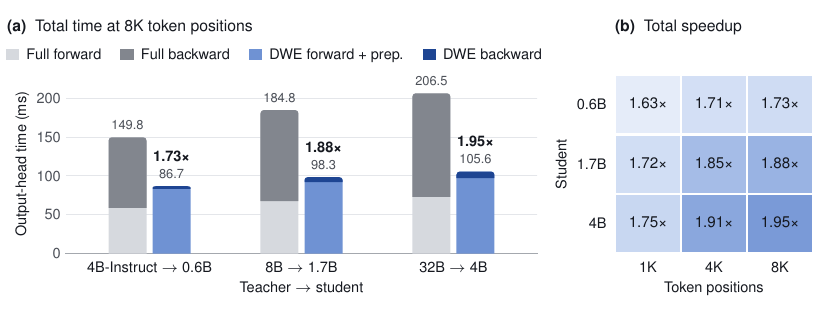}
  \caption{\textbf{Output-head time across model scales.}
  (a) We compare mean total output-head time for Full and DWE at 8K token
  positions across the three model configurations. Stacked segments show
  forward and backward time; DWE forward includes estimator preparation.
  (b) Each cell reports DWE's total output-head speedup over Full for the
  corresponding model configuration and number of token positions.
}
  \label{fig:kd-head-efficiency}
\end{figure}

We measure output-head computation for Qwen3-0.6B~\citep{yang2025qwen3},
Qwen3-1.7B~\citep{yang2025qwen3}, and Qwen3-4B~\citep{yang2025qwen3}
using fixed KD inputs. Within each configuration, Full and DWE use the
same hidden states, output weights, and numerical settings. Total time
includes the teacher and student output projections, loss computation,
estimator preparation, and backpropagation. We perform 30 warmup
iterations followed by 100 timed repetitions, interleaving the methods
within each repetition.

Figure~\ref{fig:kd-head-efficiency} shows that DWE accelerates total
output-head computation across all three model configurations,
achieving \KDHeadSpeedMin--\KDHeadSpeedMax$\times$ speedups.
These speedups arise because the added forward cost is much smaller
than the time saved in backpropagation.
For Qwen3-1.7B~\citep{yang2025qwen3} at 8K positions, DWE adds
24.3\,ms to forward computation while saving 110.8\,ms in backpropagation.

\FloatBarrier

\subsection{Training Efficiency}
\label{sec:kd-training-efficiency}

We measure KD training time to assess DWE's end-to-end efficiency.
Following the setup in Section~\ref{sec:experimental-setup}, we precompute
teacher hidden states for reuse across methods, then time student
training on the same 80M tokens.

\begin{table}[!ht]
  \centering
  \caption{\textbf{KD training efficiency.} We train each student on 80M
  tokens using its corresponding teacher. Speedup is relative to Full.}
  \label{tab:kd-training-efficiency}
  \dwetable
  \begin{tabular}{l ccr ccr ccr}
    \toprule
    & \multicolumn{3}{c}{4B-Instruct $\to$ 0.6B}
    & \multicolumn{3}{c}{8B $\to$ 1.7B}
    & \multicolumn{3}{c}{32B $\to$ 4B} \dwehs\\
    \cmidrule(lr){2-4}\cmidrule(lr){5-7}\cmidrule(lr){8-10}
    Method & GPU h & Speedup & Tokens/s & GPU h & Speedup & Tokens/s
      & GPU h & Speedup & Tokens/s \dwehs\\
    \midrule
Full & 1.09 & 1.00$\times$ & 20,460 & 1.54 & 1.00$\times$ & 14,429 & 2.75 & 1.00$\times$ & 8,077 \\
\dweours DWE & 0.92 & 1.18$\times$ & 24,131 & 1.30 & 1.19$\times$ & 17,109 & 2.47 & 1.12$\times$ & 9,011 \\
    \bottomrule
  \end{tabular}
\end{table}

Table~\ref{tab:kd-training-efficiency} shows that DWE consistently
improves end-to-end KD training efficiency across all three model
configurations. Compared with Full, DWE completes the same 80M-token
training workload with \KDScaleTrainingReduction--\KDTrainingReduction\%
less GPU time, while increasing token throughput by 11.6--18.6\%.
For Qwen3-1.7B~\citep{yang2025qwen3}, training time decreases from
\KDFullTrainingHours{} to \KDDweTrainingHours{} GPU h, yielding a
1.19$\times$ speedup. Throughput rises from 14,429 to 17,109 tokens/s
in the same configuration. These results demonstrate that the output-head
savings carry over to complete training runs, delivering consistent
reductions in GPU training cost across the evaluated model scales
(Appendix~\ref{app:runtime-analysis}).

\subsection{Downstream Performance}
\label{sec:kd-downstream}

We evaluate whether DWE's sparse gradient estimates preserve distillation
quality on seven zero-shot benchmarks. Table~\ref{tab:kd-benchmarks}
compares Full KD, IS, and DWE across the three model configurations.

\begin{table}[!ht]
  \centering
  \caption{\textbf{Downstream performance across model sizes.}
  Zero-shot benchmark scores (\%).
  Best scores among Full KD, IS, and DWE are bolded within each model group.}
  \label{tab:kd-benchmarks}
  \dwetable
  \begin{tabular}{l cccccccc}
    \toprule
    Method & HellaSwag & PIQA & ARC-E & ARC-C & GSM8K
      & LAMBADA & MMLU & Avg. \dwehs\\
    \midrule
\multicolumn{9}{l}{{\itshape\color{inkmuted}Qwen3-4B-Instruct-2507 $\to$ Qwen3-0.6B}} \\
Full KD & 49.4 & \textbf{68.3} & \textbf{66.8} & \textbf{36.0} & \textbf{11.0} & 46.0 & \textbf{45.6} & \textbf{46.2} \\
IS & \textbf{49.5} & \textbf{68.3} & \textbf{66.8} & 35.8 & 10.9 & 46.2 & 45.4 & 46.1 \\
\dweours DWE & \textbf{49.5} & 68.1 & \textbf{66.8} & 35.9 & 10.5 & \textbf{46.3} & 45.4 & 46.1 \\
\addlinespace[3pt]
\multicolumn{9}{l}{{\itshape\color{inkmuted}Qwen3-8B $\to$ Qwen3-1.7B}} \\
Full KD & 62.8 & \textbf{73.6} & 75.0 & 45.1 & 45.0 & \textbf{55.6} & 58.0 & 59.3 \\
IS & 62.8 & 73.4 & 74.8 & \textbf{45.6} & 44.8 & 55.5 & \textbf{58.2} & 59.3 \\
\dweours DWE & \textbf{62.9} & 73.2 & \textbf{75.1} & 45.4 & \textbf{45.3} & \textbf{55.6} & 58.1 & \textbf{59.4} \\
\addlinespace[3pt]
\multicolumn{9}{l}{{\itshape\color{inkmuted}Qwen3-32B $\to$ Qwen3-4B}} \\
Full KD & 71.6 & 76.6 & 82.5 & \textbf{54.9} & 62.1 & \textbf{61.6} & \textbf{69.5} & 68.4 \\
IS & 71.6 & \textbf{76.7} & \textbf{82.6} & \textbf{54.9} & \textbf{63.9} & \textbf{61.6} & \textbf{69.5} & \textbf{68.7} \\
\dweours DWE & \textbf{71.7} & 76.6 & \textbf{82.6} & \textbf{54.9} & 63.4 & 61.4 & 69.4 & 68.6 \\
    \bottomrule
  \end{tabular}
  \dwenote{Avg. is the unweighted mean of the seven displayed benchmark scores.}
\end{table}

DWE achieves average performance comparable to Full KD and IS across all
three configurations. Its average scores across the seven benchmarks are
46.1, 59.4, and 68.6 for the 0.6B, 1.7B, and 4B students, respectively.
At 1.7B, DWE matches or exceeds Full KD on six of seven benchmarks and
achieves the highest average score (59.4, compared with 59.3 for both
baselines). At 4B, it matches or exceeds Full KD on five benchmarks,
including GSM8K, where it scores 63.4 compared with 62.1 for Full KD.
These results show that sparse output-head backpropagation preserves
downstream performance across model scales and tasks. Consistent with
these findings, DWE and Full KD also yield nearly identical forward KL
divergence (0.3950 vs.\ 0.3947) and next-token perplexity
(10.508 vs.\ 10.505) on held-out text (Appendix~\ref{app:kd-heldout}).

\FloatBarrier
\newpage

\subsection{Ablation Study}
\label{sec:kd-sampling-budget}

We evaluate the contribution of each DWE component to gradient
estimation accuracy. We then compare DWE with IS at matched
sampling budgets.
Both analyses use 1,024 fixed KD token positions and measure relative
MSE against the exact hidden-state and output-weight gradients.
Relative MSE divides the squared Frobenius error by the squared Frobenius
norm of the exact gradient. Results are averaged over 128 sampling repetitions.

\begin{table}[!ht]
  \centering
  \caption{\textbf{Component ablation.} Cumulative additions at $K=8$.
  Hidden and Head report relative MSE for gradients with respect to $h$ and $W$.}
  \label{tab:kd-component-ablation}
  \dwetable
  \begin{tabular}{l l c@{\hspace{7pt}}l c@{\hspace{7pt}}l}
    \toprule
    ID & Configuration & \multicolumn{2}{l}{\hspace{.45em}Hidden}
      & \multicolumn{2}{l}{\hspace{.7em}Head} \dwehs\\
    \midrule
A0 & Importance sampling & 0.3042 & \dwebar[dwebase]{1000} & 0.1679 & \dwebar[dwebase]{1000} \\
A1 & Independent positive/negative pairs & 0.2019 & \dwebar[dwesoft]{664} & 0.1293 & \dwebar[dwesoft]{770} \\
A2 & Geometry-aware IID pairs & 0.1941 & \dwebar[dwesoft]{638} & 0.1283 & \dwebar[dwesoft]{764} \\
\dweours A3 & Geometry-aware stratified pairs (DWE) & \textbf{0.1110} & \dwebar[dwemain]{365} & \textbf{0.0707} & \dwebar[dwemain]{421} \\
    \bottomrule
  \end{tabular}
\end{table}

\begingroup
\noindent\begin{minipage}[t]{.53\textwidth}
\vspace{0pt}
\raggedright
\paragraph{\mbox{Paired construction.}}
A1 samples eight positive and eight negative coordinates independently,
without geometry-aware pairing or stratification. Compared with
IS at the same 16-draw budget, pairing reduces hidden-state and output-weight
gradient errors by \KDPairHiddenReduction\% and \KDPairHeadReduction\%
(A0 vs.\ A1).

\paragraph{\mbox{Coupling and stratification.}}
Geometry-aware pairing reduces hidden-state gradient error by
\KDGeometryHiddenReduction\% under IID sampling (A1 vs.\ A2).
Adding stratification reduces hidden-state and output-weight gradient
errors by a further
\KDStratificationHiddenReduction\% and \KDStratificationHeadReduction\%
(A2 vs.\ A3), yielding the final DWE estimator. Relative to IS,
DWE reduces hidden-state and output-weight gradient MSE by
\KDOverallHiddenReduction\% and \KDOverallHeadReduction\%, respectively
(A0 vs.\ A3, Table~\ref{tab:kd-component-ablation}).

\paragraph{Sampling budget.}
Figure~\ref{fig:kd-sampling-budget} shows that DWE achieves lower
gradient estimation error than IS across all evaluated sampling budgets.
At comparable hidden-state gradient error, IS requires 44 coordinate
draws per token, compared with 16 for DWE, using $2.75\times$ as many draws.
As $K$ increases from 1 to 32, DWE's total output-head time on 1K
positions rises from \KDBudgetTimeFirst{} to \KDBudgetTimeLast{}\,ms,
remaining below Full's \KDBudgetTimeFull{}\,ms. We use $K=8$ in the
main training comparisons.

\end{minipage}\hfill
\begin{minipage}[t]{.44\textwidth}
\vspace{0pt}
  \centering
  \includegraphics[width=\linewidth]{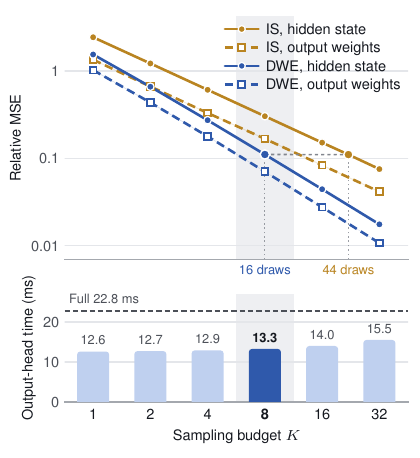}\par
  \setlength{\abovecaptionskip}{4pt}
  \expandafter\def\csname @captype\endcsname{figure}\caption[Gradient estimation error and output-head time.]{\raggedright
  \textbf{Gradient estimation error and output-head time.}
  Top: relative MSE with 95\% Monte Carlo confidence intervals.
  Bottom: DWE runtime on 1K positions, with Full as reference.
  Both methods use $2K$ draws per token.}
  \label{fig:kd-sampling-budget}
\end{minipage}
\par
\endgroup
\FloatBarrier
\section{Related Work}
\label{sec:related-work}

\paragraph{Efficient distillation.}
Prior work on efficient distillation focuses on reducing the cost of
teacher supervision through top-$k$ selection and token sampling.
Top-$k$ methods~\citep{peng2025pretraining,ryskulov2026efficient,
li2026rethinking} retain high-probability tokens, while
TA-OPD~\citep{huang2026taopd} aggregates
the remaining probability mass into a tail category.
Sampling-based methods~\citep{anshumann2025sparse,ko2026reopold,
xing2026tropd,oh2026vopd} also estimate distillation signals
using teacher feedback on sampled tokens.
\method{} further improves the efficiency of gradient propagation
through the output head.

\paragraph{Efficient backpropagation.}
Prior work has explored sparse backpropagation to improve training
efficiency. meProp~\citep{sun2017meprop} retains large-magnitude
gradient coordinates, producing sparse but biased approximations.
IS~\citep{wang2018atomo} instead samples and reweights gradient
contributions to obtain unbiased estimates.
Randomized automatic differentiation~\citep{oktay2021randomized}
extends unbiased estimation to computational graphs.
Unbiased vector-Jacobian product
approximations~\citep{bakong2026unbiased} further analyze variance
propagation and derive estimators under sparsity constraints.
\method{} retains unbiasedness while further reducing gradient
estimation variance through zero-sum pairing, geometry-aware
coupling, and stratified sampling.

\paragraph{Concurrent work.}
Concurrent with our work, SparseOPD~\citep{miao2026sparseopd} explored
vocabulary sparsification for output-head backpropagation in on-policy
distillation. It constructs the full logit gradient
before backpropagating through selected output rows. It selects
large-magnitude positive and negative entries and redistributes discarded
mass. The resulting deterministic approximation preserves the signed
mass but is generally biased. \method{} constructs randomized paired
estimates that recover the full gradient in expectation.
\section*{Discussion and Limitations}

\paragraph{Applicability.}
Although we evaluate \method{} in distillation, its construction requires
only a zero-sum logit gradient that can be computed exactly at each
position. It therefore applies more broadly to differentiable objectives
that match a softmax output to a fixed target distribution.
In soft-target distillation, both the positive and negative parts of the
gradient can span multiple tokens. With one-hot cross-entropy, the negative
part is confined to the label, so every pair contains that fixed label and
one sampled non-label coordinate.

\paragraph{Sampling variance.}
DWE preserves the full-vocabulary gradient in expectation, while sampling
noise can lead to different optimization trajectories. Increasing $K$
reduces the estimation variance at the cost of more output-head backward
computation, making the sampling budget a trade-off between estimation
accuracy and computational savings.

\paragraph{End-to-end gains.}
\method{} reduces computation only in the output-head backward pass.
End-to-end gains therefore depend on this pass's share of training cost. They can be
larger for small students with large vocabularies, and smaller when rollout
generation or teacher inference dominates each step, as in OPD.
\section{Conclusion}
\label{sec:conclusion}

In this work, we introduced Distillation with Errors (\method{}), a method
for efficient output-head backpropagation in knowledge distillation.
\method{} uses paired sampling to construct sparse, unbiased estimates
of the full logit gradient. Our off-policy KD experiments demonstrate reduced
output-head computation and training time, with lower gradient estimation
error than importance sampling at the same sampling budget. Additional
OPD experiments show performance comparable to full-vocabulary distillation.
\setlength{\parskip}{6pt}
\setlength{\textfloatsep}{20pt plus 2pt minus 4pt}
\setlength{\intextsep}{12pt plus 2pt minus 2pt}
\bibliographystyle{iclr2027_conference}
\bibliography{references}

@inproceedings{wang2018atomo,
  title = {{ATOMO}: Communication-efficient Learning via Atomic Sparsification},
  author = {Wang, Hongyi and Sievert, Scott and Liu, Shengchao and Charles, Zachary and Papailiopoulos, Dimitris and Wright, Stephen},
  booktitle = {Advances in Neural Information Processing Systems},
  volume = {31},
  year = {2018},
  url = {https://proceedings.neurips.cc/paper/2018/hash/33b3214d792caf311e1f00fd22b392c5-Abstract.html}
}

@inproceedings{douc2005comparison,
  title = {Comparison of Resampling Schemes for Particle Filtering},
  author = {Douc, Randal and Capp{\'e}, Olivier and Moulines, Eric},
  booktitle = {Proceedings of the 4th International Symposium on Image and Signal Processing and Analysis},
  pages = {64--69},
  year = {2005},
  url = {https://arxiv.org/abs/cs/0507025}
}

@inproceedings{agarwal2024gkd,
  title = {On-Policy Distillation of Language Models: Learning from Self-Generated Mistakes},
  author = {Agarwal, Rishabh and Vieillard, Nino and Zhou, Yongchao and Stanczyk, Piotr and Ramos Garea, Sabela and Geist, Matthieu and Bachem, Olivier},
  booktitle = {International Conference on Learning Representations},
  year = {2024},
  url = {https://proceedings.iclr.cc/paper_files/paper/2024/hash/5be69a584901a26c521c2b51e40a4c20-Abstract-Conference.html}
}

@inproceedings{gu2024minillm,
  title = {{MiniLLM}: Knowledge Distillation of Large Language Models},
  author = {Gu, Yuxian and Dong, Li and Wei, Furu and Huang, Minlie},
  booktitle = {International Conference on Learning Representations},
  year = {2024},
  url = {https://proceedings.iclr.cc/paper_files/paper/2024/hash/8ac015d409635f196f9e3e9dcfb9a94e-Abstract-Conference.html}
}

@article{li2026rethinking,
  title = {Rethinking On-Policy Distillation of Large Language Models: Phenomenology, Mechanism, and Recipe},
  author = {Li, Yaxuan and Zuo, Yuxin and He, Bingxiang and Zhang, Jinqian and Xiao, Chaojun and Qian, Cheng and Yu, Tianyu and Gao, Huan-ang and Yang, Wenkai and Liu, Zhiyuan and Ding, Ning},
  journal = {arXiv preprint arXiv:2604.13016},
  year = {2026},
  doi = {10.48550/arXiv.2604.13016},
  url = {https://arxiv.org/abs/2604.13016v2}
}

@inproceedings{anshumann2025sparse,
  title = {Sparse Logit Sampling: Accelerating Knowledge Distillation in {LLMs}},
  author = {Anshumann and Zaidi, Mohd Abbas and Kedia, Akhil and Ahn, Jinwoo and Kwon, Taehwak and Lee, Kangwook and Lee, Haejun and Lee, Joohyung},
  booktitle = {Proceedings of the 63rd Annual Meeting of the Association for Computational Linguistics (Volume 1: Long Papers)},
  year = {2025},
  month = jul,
  pages = {18085--18108},
  publisher = {Association for Computational Linguistics},
  address = {Vienna, Austria},
  doi = {10.18653/v1/2025.acl-long.885},
  url = {https://aclanthology.org/2025.acl-long.885/}
}

@article{yang2025qwen3,
  title = {{Qwen3} Technical Report},
  author = {Yang, An and Li, Anfeng and Yang, Baosong and Zhang, Beichen and Hui, Binyuan and Zheng, Bo and Yu, Bowen and Gao, Chang and Huang, Chengen and Lv, Chenxu and Zheng, Chujie and Liu, Dayiheng and Zhou, Fan and Huang, Fei and Hu, Feng and Ge, Hao and Wei, Haoran and Lin, Huan and Tang, Jialong and Yang, Jian and Tu, Jianhong and Zhang, Jianwei and Yang, Jianxin and Yang, Jiaxi and Zhou, Jing and Zhou, Jingren and Lin, Junyang and Dang, Kai and Bao, Keqin and Yang, Kexin and Yu, Le and Deng, Lianghao and Li, Mei and Xue, Mingfeng and Li, Mingze and Zhang, Pei and Wang, Peng and Zhu, Qin and Men, Rui and Gao, Ruize and Liu, Shixuan and Luo, Shuang and Li, Tianhao and Tang, Tianyi and Yin, Wenbiao and Ren, Xingzhang and Wang, Xinyu and Zhang, Xinyu and Ren, Xuancheng and Fan, Yang and Su, Yang and Zhang, Yichang and Zhang, Yinger and Wan, Yu and Liu, Yuqiong and Wang, Zekun and Cui, Zeyu and Zhang, Zhenru and Zhou, Zhipeng and Qiu, Zihan},
  journal = {arXiv preprint arXiv:2505.09388},
  year = {2025},
  url = {https://arxiv.org/abs/2505.09388}
}

@article{heo2026opd2,
  title = {On-Policy Delta Distillation},
  author = {Heo, Byeongho and Hwang, Jaehui and Yun, Sangdoo and Han, Dongyoon},
  journal = {arXiv preprint arXiv:2607.15161},
  year = {2026},
  url = {https://arxiv.org/abs/2607.15161}
}

@article{moshkov2025aimo2,
  title = {{AIMO-2} Winning Solution: Building State-of-the-Art Mathematical Reasoning Models with {OpenMathReasoning} dataset},
  author = {Moshkov, Ivan and Hanley, Darragh and Sorokin, Ivan and Toshniwal, Shubham and Henkel, Christof and Schifferer, Benedikt and Du, Wei and Gitman, Igor},
  journal = {arXiv preprint arXiv:2504.16891},
  year = {2025},
  url = {https://arxiv.org/abs/2504.16891}
}

@article{ahmad2025opencode,
  title = {{OpenCodeReasoning}: Advancing Data Distillation for Competitive Coding},
  author = {Ahmad, Wasi Uddin and Narenthiran, Sean and Majumdar, Somshubra and Ficek, Aleksander and Jain, Siddhartha and Huang, Jocelyn and Noroozi, Vahid and Ginsburg, Boris},
  journal = {arXiv preprint arXiv:2504.01943},
  year = {2025},
  url = {https://arxiv.org/abs/2504.01943}
}

@inproceedings{hendrycks2021math,
  author = {Hendrycks, Dan and Burns, Collin and Kadavath, Saurav and Arora, Akul and Basart, Steven and Tang, Eric and Song, Dawn and Steinhardt, Jacob},
  booktitle = {Proceedings of the Neural Information Processing Systems Track on Datasets and Benchmarks},
  title = {Measuring Mathematical Problem Solving With the {MATH} Dataset},
  url = {https://datasets-benchmarks-proceedings.neurips.cc/paper_files/paper/2021/file/be83ab3ecd0db773eb2dc1b0a17836a1-Paper-round2.pdf},
  volume = {1},
  year = {2021},
}

@inproceedings{lightman2023verify,
  author = {Lightman, Hunter and Kosaraju, Vineet and Burda, Yuri and Edwards, Harrison and Baker, Bowen and Lee, Teddy and Leike, Jan and Schulman, John and Sutskever, Ilya and Cobbe, Karl},
  booktitle = {International Conference on Learning Representations},
  pages = {39578--39601},
  title = {Let's Verify Step by Step},
  url = {https://proceedings.iclr.cc/paper_files/paper/2024/file/aca97732e30bcf1303bc22ac3924fd16-Paper-Conference.pdf},
  volume = {2024},
  year = {2024},
}

@inproceedings{he2024olympiad,
  title = {{O}lympiad{B}ench: A Challenging Benchmark for Promoting {AGI} with Olympiad-Level Bilingual Multimodal Scientific Problems},
  author = {He, Chaoqun and Luo, Renjie and Bai, Yuzhuo and Hu, Shengding and Thai, Zhen and Shen, Junhao and Hu, Jinyi and Han, Xu and Huang, Yujie and Zhang, Yuxiang and Liu, Jie and Qi, Lei and Liu, Zhiyuan and Sun, Maosong},
  booktitle = {Proceedings of the 62nd Annual Meeting of the {A}ssociation for {C}omputational {L}inguistics (Volume 1: Long Papers)},
  year = {2024},
  address = {Bangkok, Thailand},
  publisher = {Association for Computational Linguistics},
  url = {https://aclanthology.org/2024.acl-long.211/},
  doi = {10.18653/v1/2024.acl-long.211},
  pages = {3828--3850},
}

@inproceedings{stojanovski2025gym,
 author = {Stojanovski, Zafir and Stanley, Oliver and Sharratt, Joe and Jones, Richard and Adefioye, Abdulhakeem and Kaddour, Jean and K\"{o}pf, Andreas},
 booktitle = {Advances in Neural Information Processing Systems},
 doi = {10.52202/085713-1913},
 editor = {D. Belgrave and C. Zhang and H. Lin and R. Pascanu and P. Koniusz and M. Ghassemi and N. Chen},
 pages = {},
 publisher = {Curran Associates, Inc.},
 title = {Reasoning Gym: Reasoning Environments for Reinforcement Learning with Verifiable Rewards},
 url = {https://proceedings.neurips.cc/paper_files/paper/2025/file/528d56195a2c77c808494c86fa7c77ad-Paper-Datasets_and_Benchmarks_Track.pdf},
 volume = {38, Main Conference},
 year = {2025}
}

@article{li2022alphacode,
  title = {Competition-level code generation with {AlphaCode}},
  volume = {378},
  url = {https://doi.org/10.1126/science.abq1158},
  doi = {10.1126/science.abq1158},
  number = {6624},
  journal = {Science},
  publisher = {American Association for the Advancement of Science (AAAS)},
  author = {Li, Yujia and Choi, David and Chung, Junyoung and Kushman, Nate and Schrittwieser, Julian and Leblond, Rémi and Eccles, Tom and Keeling, James and Gimeno, Felix and Dal Lago, Agustin and Hubert, Thomas and Choy, Peter and de Masson d’Autume, Cyprien and Babuschkin, Igor and Chen, Xinyun and Huang, Po-Sen and Welbl, Johannes and Gowal, Sven and Cherepanov, Alexey and Molloy, James and Mankowitz, Daniel J. and Sutherland Robson, Esme and Kohli, Pushmeet and de Freitas, Nando and Kavukcuoglu, Koray and Vinyals, Oriol},
  year = {2022},
  pages = {1092--1097},
}

@inproceedings{jain2024livecode,
  author = {Jain, Naman and Han, King and Gu, Alex and Li, Wen-Ding and Yan, Fanjia and Zhang, Tianjun and Wang, Sida I. and Solar-Lezama, Armando and Sen, Koushik and Stoica, Ion},
  booktitle = {International Conference on Learning Representations},
  pages = {58791--58831},
  title = {{LiveCodeBench}: Holistic and Contamination Free Evaluation of Large Language Models for Code},
  url = {https://proceedings.iclr.cc/paper_files/paper/2025/file/94074dd5a072d28ff75a76dabed43767-Paper-Conference.pdf},
  volume = {2025},
  year = {2025},
}

@inproceedings{rein2023gpqa,
      title={{GPQA}: A Graduate-Level Google-Proof Q\&A Benchmark},
      author={David Rein and Betty Li Hou and Asa Cooper Stickland and Jackson Petty and Richard Yuanzhe Pang and Julien Dirani and Julian Michael and Samuel R. Bowman},
      booktitle={First Conference on Language Modeling},
      year={2024},
      url={https://openreview.net/forum?id=Ti67584b98}
}

@misc{team2025supergpqa,
      title={SuperGPQA: Scaling LLM Evaluation across 285 Graduate Disciplines}, 
      author={{M-A-P Team} and Xinrun Du and Yifan Yao and Kaijing Ma and Bingli Wang and Tianyu Zheng and Kang Zhu and Minghao Liu and Yiming Liang and Xiaolong Jin and Zhenlin Wei and Chujie Zheng and Kaixing Deng and Shuyue Guo and Shian Jia and Sichao Jiang and Yiyan Liao and Rui Li and Qinrui Li and Sirun Li and Yizhi Li and Yunwen Li and Dehua Ma and Yuansheng Ni and Haoran Que and Qiyao Wang and Zhoufutu Wen and Siwei Wu and Tianshun Xing and Ming Xu and Zhenzhu Yang and Zekun Moore Wang and Junting Zhou and Yuelin Bai and Xingyuan Bu and Chenglin Cai and Liang Chen and Yifan Chen and Chengtuo Cheng and Tianhao Cheng and Keyi Ding and Siming Huang and Yun Huang and Yaoru Li and Yizhe Li and Zhaoqun Li and Tianhao Liang and Chengdong Lin and Hongquan Lin and Yinghao Ma and Zhongyuan Peng and Zifan Peng and Qige Qi and Shi Qiu and Xingwei Qu and Yizhou Tan and Zili Wang and Chenqing Wang and Hao Wang and Yiya Wang and Yubo Wang and Jiajun Xu and Kexin Yang and Ruibin Yuan and Yuanhao Yue and Tianyang Zhan and Chun Zhang and Jingyang Zhang and Xiyue Zhang and Xingjian Zhang and Yue Zhang and Yongchi Zhao and Xiangyu Zheng and Chenghua Zhong and Yang Gao and Zhoujun Li and Dayiheng Liu and Qian Liu and Tianyu Liu and Shiwen Ni and Junran Peng and Yujia Qin and Wenbo Su and Guoyin Wang and Shi Wang and Jian Yang and Min Yang and Meng Cao and Xiang Yue and Zhaoxiang Zhang and Wangchunshu Zhou and Jiaheng Liu and Qunshu Lin and Wenhao Huang and Ge Zhang},
      year={2025},
      eprint={2502.14739},
      archivePrefix={arXiv},
      primaryClass={cs.CL},
      url={https://arxiv.org/abs/2502.14739}, 
}

@inproceedings{wang2023scibench,
  title = {{S}ci{B}ench: Evaluating College-Level Scientific Problem-Solving Abilities of Large Language Models},
  author = {Wang, Xiaoxuan and Hu, Ziniu and Lu, Pan and Zhu, Yanqiao and Zhang, Jieyu and Subramaniam, Satyen and Loomba, Arjun R and Zhang, Shichang and Sun, Yizhou and Wang, Wei},
  booktitle = {Proceedings of the 41st International Conference on Machine Learning},
  pages = {50622--50649},
  year = {2024},
  volume = {235},
  series = {Proceedings of Machine Learning Research},
  publisher = {PMLR},
  url = {https://proceedings.mlr.press/v235/wang24z.html},
}

@misc{nvidia2025science,
  author = {{NVIDIA}},
  title = {{OpenScienceReasoning-2}},
  year = {2025},
  url = {https://huggingface.co/datasets/nvidia/OpenScienceReasoning-2}
}

@misc{maa2025aime,
  author = {{MAA}},
  organization = {Mathematical Association of America},
  title = {{American Invitational Mathematics Examination (AIME), 2025}},
  year = {2025},
  url = {https://maa.org/maa-invitational-competitions/}
}

@misc{maa2023amc,
  author = {{MAA}},
  organization = {Mathematical Association of America},
  title = {{American Mathematics Competitions (AMC), 2023}},
  year = {2023},
  url = {https://maa.org/student-programs/amc/}
}

@article{matharena2025hmmt,
      title={Beyond Benchmarks: MathArena as an Evaluation Platform for Mathematics with LLMs}, 
      author={Jasper Dekoninck and Nikola Jovanović and Tim Gehrunger and Kári Rögnvaldsson and Ivo Petrov and Chenhao Sun and Martin Vechev},
      year={2026},
      eprint={2605.00674},
      archivePrefix={arXiv},
      primaryClass={cs.CL},
      url={https://arxiv.org/abs/2605.00674}, 
}

@misc{openr12025codeforces,
  author = {Penedo, Guilherme and Lozhkov, Anton and Kydl{\'i}{\v c}ek, Hynek and Ben Allal, Loubna and Beeching, Edward and Piqueres Lajar{\'i}n, Agust{\'i}n and Gallou{\'e}dec, Quentin and Habib, Nathan and Tunstall, Lewis and von Werra, Leandro},
  title = {{CodeForces}},
  year = {2025},
  publisher = {Hugging Face},
  url = {https://huggingface.co/datasets/open-r1/codeforces}
}

@article{peyre2019computational,
  title = {Computational Optimal Transport},
  author = {Peyr{\'e}, Gabriel and Cuturi, Marco},
  journal = {Foundations and Trends in Machine Learning},
  volume = {11},
  number = {5--6},
  pages = {355--607},
  year = {2019},
  doi = {10.1561/2200000073},
  url = {https://doi.org/10.1561/2200000073}
}

@article{huang2026taopd,
  title = {{Tail-Aware Top-$k$ On-Policy Distillation}},
  author = {Huang, Huipeng and Wei, Hongxin},
  journal = {arXiv preprint arXiv:2608.14728},
  year = {2026},
  url = {https://arxiv.org/abs/2608.14728}
}

@article{hinton2015distilling,
  title = {Distilling the Knowledge in a Neural Network},
  author = {Hinton, Geoffrey and Vinyals, Oriol and Dean, Jeff},
  journal = {arXiv preprint arXiv:1503.02531},
  year = {2015},
  url = {https://arxiv.org/abs/1503.02531}
}

@article{sanh2019distilbert,
  title = {{DistilBERT}, a Distilled Version of {BERT}: Smaller, Faster, Cheaper and Lighter},
  author = {Sanh, Victor and Debut, Lysandre and Chaumond, Julien and Wolf, Thomas},
  journal = {arXiv preprint arXiv:1910.01108},
  year = {2019},
  url = {https://arxiv.org/abs/1910.01108}
}

@inproceedings{sun2017meprop,
  title = {me{P}rop: Sparsified Back Propagation for Accelerated Deep Learning with Reduced Overfitting},
  author = {Sun, Xu and Ren, Xuancheng and Ma, Shuming and Wang, Houfeng},
  booktitle = {Proceedings of the 34th International Conference on Machine Learning},
  series = {Proceedings of Machine Learning Research},
  volume = {70},
  pages = {3299--3308},
  year = {2017},
  publisher = {PMLR},
  url = {https://proceedings.mlr.press/v70/sun17c.html}
}

@inproceedings{peng2025pretraining,
  title = {Pre-training Distillation for Large Language Models: A Design Space Exploration},
  author = {Peng, Hao and Lv, Xin and Bai, Yushi and Yao, Zijun and Zhang, Jiajie and Hou, Lei and Li, Juanzi},
  editor = {Che, Wanxiang and Nabende, Joyce and Shutova, Ekaterina and Pilehvar, Mohammad Taher},
  booktitle = {Proceedings of the 63rd Annual Meeting of the Association for Computational Linguistics (Volume 1: Long Papers)},
  month = jul,
  year = {2025},
  address = {Vienna, Austria},
  publisher = {Association for Computational Linguistics},
  url = {https://aclanthology.org/2025.acl-long.181/},
  doi = {10.18653/v1/2025.acl-long.181},
  pages = {3603--3618},
  ISBN = {979-8-89176-251-0}
}

@misc{ryskulov2026efficient,
  title = {Efficient Knowledge Distillation for {LLMs}: Offline Top-{K} Logits and a Fused Chunked {KL} Loss},
  author = {Bakbergen Ryskulov and Iker Garc{\'i}a-Ferrero and David Montero and David Jansen and Ali Hashemi and Jezabel R. Garcia and Antonio Tiene and Rom{\'a}n Or{\'u}s},
  year = {2026},
  eprint = {2608.03796},
  archivePrefix = {arXiv},
  url = {https://arxiv.org/abs/2608.03796}
}

@misc{ko2026reopold,
  title = {Scaling Reasoning Efficiently via Relaxed On-Policy Distillation},
  author = {Jongwoo Ko and Sara Abdali and Young Jin Kim and Tianyi Chen and Pashmina Cameron},
  year = {2026},
  eprint = {2603.11137},
  archivePrefix = {arXiv},
  primaryClass = {cs.LG},
  url = {https://arxiv.org/abs/2603.11137}
}

@misc{xing2026tropd,
  title = {Trust Region On-Policy Distillation},
  author = {Xingrun Xing and Haoqing Wang and Boyan Gao and Ziheng Li and Yehui Tang},
  year = {2026},
  eprint = {2606.01249},
  archivePrefix = {arXiv},
  primaryClass = {cs.LG},
  url = {https://arxiv.org/abs/2606.01249}
}

@misc{oh2026vopd,
  title = {{KL} for a {KL}: On-Policy Distillation with Control Variate Baseline},
  author = {Minjae Oh and Sangjun Song and Gyubin Choi and Yunho Choi and Yohan Jo},
  year = {2026},
  eprint = {2605.07865},
  archivePrefix = {arXiv},
  url = {https://arxiv.org/abs/2605.07865}
}

@misc{miao2026sparseopd,
  title = {Look Before You Select: Rethinking Vocabulary Sparsification in On-Policy Distillation},
  author = {Yongliang Miao and Shuang Liu and Yanguang Liu and Yandong Bai and Mengnan Du},
  year = {2026},
  eprint = {2609.34386},
  archivePrefix = {arXiv},
  url = {https://arxiv.org/abs/2609.34386}
}

@inproceedings{penedo2024fineweb,
  title={The {FineWeb} Datasets: Decanting the Web for the Finest Text Data at Scale},
  author={Penedo, Guilherme and Kydl{\'i}{\v c}ek, Hynek and Ben Allal, Loubna and Lozhkov, Anton and Mitchell, Margaret and Raffel, Colin and Von Werra, Leandro and Wolf, Thomas},
  booktitle={Advances in Neural Information Processing Systems},
  volume={37},
  pages={30811--30849},
  publisher={Curran Associates, Inc.},
  year={2024},
  doi={10.52202/079017-0970},
  url={https://proceedings.neurips.cc/paper_files/paper/2024/hash/370df50ccfdf8bde18f8f9c2d9151bda-Abstract-Datasets_and_Benchmarks_Track.html}
}

@inproceedings{zellers2019hellaswag,
  title={{HellaSwag}: Can a Machine Really Finish Your Sentence?},
  author={Zellers, Rowan and Holtzman, Ari and Bisk, Yonatan and Farhadi, Ali and Choi, Yejin},
  booktitle={Proceedings of the 57th Annual Meeting of the Association for Computational Linguistics},
  pages={4791--4800},
  year={2019},
  doi={10.18653/v1/P19-1472},
  url={https://aclanthology.org/P19-1472/}
}

@inproceedings{bisk2020piqa,
  title={{PIQA}: Reasoning about Physical Commonsense in Natural Language},
  author={Bisk, Yonatan and Zellers, Rowan and Le Bras, Ronan and Gao, Jianfeng and Choi, Yejin},
  booktitle={Proceedings of the AAAI Conference on Artificial Intelligence},
  volume={34},
  pages={7432--7439},
  year={2020},
  doi={10.1609/aaai.v34i05.6239},
  url={https://ojs.aaai.org/index.php/AAAI/article/view/6239}
}

@article{clark2018arc,
  title={Think You Have Solved Question Answering? Try {ARC}, the {AI2} Reasoning Challenge},
  author={Clark, Peter and Cowhey, Isaac and Etzioni, Oren and Khot, Tushar and Sabharwal, Ashish and Schoenick, Carissa and Tafjord, Oyvind},
  journal={arXiv preprint arXiv:1803.05457},
  year={2018},
  url={https://arxiv.org/abs/1803.05457}
}

@article{cobbe2021verifiers,
  title={Training Verifiers to Solve Math Word Problems},
  author={Cobbe, Karl and Kosaraju, Vineet and Bavarian, Mohammad and Chen, Mark and Jun, Heewoo and Kaiser, Lukasz and Plappert, Matthias and Tworek, Jerry and Hilton, Jacob and Nakano, Reiichiro and Hesse, Christopher and Schulman, John},
  journal={arXiv preprint arXiv:2110.14168},
  year={2021},
  url={https://arxiv.org/abs/2110.14168}
}

@inproceedings{paperno2016lambada,
  title={The {LAMBADA} Dataset: Word Prediction Requiring a Broad Discourse Context},
  author={Paperno, Denis and Kruszewski, Germ{\'a}n and Lazaridou, Angeliki and Pham, Ngoc Quan and Bernardi, Raffaella and Pezzelle, Sandro and Baroni, Marco and Boleda, Gemma and Fern{\'a}ndez, Raquel},
  booktitle={Proceedings of the 54th Annual Meeting of the Association for Computational Linguistics (Volume 1: Long Papers)},
  pages={1525--1534},
  year={2016},
  doi={10.18653/v1/P16-1144},
  url={https://aclanthology.org/P16-1144/}
}

@inproceedings{hendrycks2021mmlu,
  title={Measuring Massive Multitask Language Understanding},
  author={Hendrycks, Dan and Burns, Collin and Basart, Steven and Zou, Andy and Mazeika, Mantas and Song, Dawn and Steinhardt, Jacob},
  booktitle={International Conference on Learning Representations},
  year={2021},
  url={https://arxiv.org/abs/2009.03300}
}

@misc{eval-harness,
  author={Gao, Leo and Tow, Jonathan and Abbasi, Baber and Biderman, Stella and Black, Sid and DiPofi, Anthony and Foster, Charles and Golding, Laurence and Hsu, Jeffrey and Le Noac'h, Alain and Li, Haonan and McDonell, Kyle and Muennighoff, Niklas and Ociepa, Chris and Phang, Jason and Reynolds, Laria and Schoelkopf, Hailey and Skowron, Aviya and Sutawika, Lintang and Tang, Eric and Thite, Anish and Wang, Ben and Wang, Kevin and Zou, Andy},
  title={The Language Model Evaluation Harness},
  year={2024},
  publisher={Zenodo},
  doi={10.5281/zenodo.12608602},
  url={https://zenodo.org/records/12608602}
}

@article{tavor2026rethinking,
  title = {Rethinking Selective Knowledge Distillation},
  author = {Tavor, Almog and Ebenspanger, Itay and Cnaan, Neil and Geva, Mor},
  journal = {arXiv preprint arXiv:2602.01395},
  year = {2026},
  doi = {10.48550/arXiv.2602.01395},
  url = {https://arxiv.org/abs/2602.01395}
}

@misc{lozhkov2024finewebedu,
  author = {Lozhkov, Anton and Ben Allal, Loubna and von Werra, Leandro and Wolf, Thomas},
  title = {{FineWeb-Edu}: the Finest Collection of Educational Content},
  year = {2024},
  publisher = {Hugging Face},
  doi = {10.57967/hf/2497},
  url = {https://huggingface.co/datasets/HuggingFaceFW/fineweb-edu}
}

@misc{guha2024evalchemy,
  author = {Guha, Etash and Raoff, Negin and Mercat, Jean and Marten, Ryan and Frankel, Eric and Keh, Sedrick and Grover, Sachin and Smyrnis, George and Vu, Trung and Saad-Falcon, Jon and Choi, Caroline and Arora, Kushal and Merrill, Mike and Deng, Yichuan and Suvarna, Ashima and Bansal, Hritik and Nezhurina, Marianna and Heckel, Reinhard and Oh, Seewong and Hashimoto, Tatsunori and Jitsev, Jenia and Choi, Yejin and Shankar, Vaishaal and Dimakis, Alex and Sathiamoorthy, Mahesh and Schmidt, Ludwig},
  title = {{Evalchemy}},
  year = {2024},
  publisher = {GitHub},
  url = {https://github.com/mlfoundations/evalchemy}
}

@inproceedings{oktay2021randomized,
  title = {Randomized Automatic Differentiation},
  author = {Oktay, Deniz and McGreivy, Nick and Aduol, Joshua and Beatson, Alex and Adams, Ryan P.},
  booktitle = {International Conference on Learning Representations},
  year = {2021},
  url = {https://lips.cs.princeton.edu/pdfs/oktay2021randomized.pdf}
}

@article{bakong2026unbiased,
  title = {Unbiased Approximate Vector-Jacobian Products for Efficient Backpropagation},
  author = {Bakong, Killian and Massouli{\'e}, Laurent and Oyallon, Edouard and Scaman, Kevin},
  journal = {arXiv preprint arXiv:2602.14701},
  year = {2026},
  url = {https://arxiv.org/abs/2602.14701}
}

@misc{grattafiori2024llama,
      title={The {Llama 3} Herd of Models},
      author={Aaron Grattafiori and Abhimanyu Dubey and Abhinav Jauhri and Abhinav Pandey and Abhishek Kadian and Ahmad Al-Dahle and Aiesha Letman and Akhil Mathur and Alan Schelten and Alex Vaughan and Amy Yang and Angela Fan and Anirudh Goyal and Anthony Hartshorn and Aobo Yang and Archi Mitra and Archie Sravankumar and Artem Korenev and Arthur Hinsvark and Arun Rao and Aston Zhang and Aurelien Rodriguez and Austen Gregerson and Ava Spataru and Baptiste Roziere and Bethany Biron and Binh Tang and Bobbie Chern and Charlotte Caucheteux and Chaya Nayak and Chloe Bi and Chris Marra and Chris McConnell and Christian Keller and Christophe Touret and Chunyang Wu and Corinne Wong and Cristian Canton Ferrer and Cyrus Nikolaidis and Damien Allonsius and Daniel Song and Danielle Pintz and Danny Livshits and Danny Wyatt and David Esiobu and Dhruv Choudhary and Dhruv Mahajan and Diego Garcia-Olano and Diego Perino and Dieuwke Hupkes and Egor Lakomkin and Ehab AlBadawy and Elina Lobanova and Emily Dinan and Eric Michael Smith and Filip Radenovic and Francisco Guzmán and Frank Zhang and Gabriel Synnaeve and Gabrielle Lee and Georgia Lewis Anderson and Govind Thattai and Graeme Nail and Gregoire Mialon and Guan Pang and Guillem Cucurell and Hailey Nguyen and Hannah Korevaar and Hu Xu and Hugo Touvron and Iliyan Zarov and Imanol Arrieta Ibarra and Isabel Kloumann and Ishan Misra and Ivan Evtimov and Jack Zhang and Jade Copet and Jaewon Lee and Jan Geffert and Jana Vranes and Jason Park and Jay Mahadeokar and Jeet Shah and Jelmer van der Linde and Jennifer Billock and Jenny Hong and Jenya Lee and Jeremy Fu and Jianfeng Chi and Jianyu Huang and Jiawen Liu and Jie Wang and Jiecao Yu and Joanna Bitton and Joe Spisak and Jongsoo Park and Joseph Rocca and Joshua Johnstun and Joshua Saxe and Junteng Jia and Kalyan Vasuden Alwala and Karthik Prasad and Kartikeya Upasani and Kate Plawiak and Ke Li and Kenneth Heafield and Kevin Stone and Khalid El-Arini and Krithika Iyer and Kshitiz Malik and Kuenley Chiu and Kunal Bhalla and Kushal Lakhotia and Lauren Rantala-Yeary and Laurens van der Maaten and Lawrence Chen and Liang Tan and Liz Jenkins and Louis Martin and Lovish Madaan and Lubo Malo and Lukas Blecher and Lukas Landzaat and Luke de Oliveira and Madeline Muzzi and Mahesh Pasupuleti and Mannat Singh and Manohar Paluri and Marcin Kardas and Maria Tsimpoukelli and Mathew Oldham and Mathieu Rita and Maya Pavlova and Melanie Kambadur and Mike Lewis and Min Si and Mitesh Kumar Singh and Mona Hassan and Naman Goyal and Narjes Torabi and Nikolay Bashlykov and Nikolay Bogoychev and Niladri Chatterji and Ning Zhang and Olivier Duchenne and Onur Çelebi and Patrick Alrassy and Pengchuan Zhang and Pengwei Li and Petar Vasic and Peter Weng and Prajjwal Bhargava and Pratik Dubal and Praveen Krishnan and Punit Singh Koura and Puxin Xu and Qing He and Qingxiao Dong and Ragavan Srinivasan and Raj Ganapathy and Ramon Calderer and Ricardo Silveira Cabral and Robert Stojnic and Roberta Raileanu and Rohan Maheswari and Rohit Girdhar and Rohit Patel and Romain Sauvestre and Ronnie Polidoro and Roshan Sumbaly and Ross Taylor and Ruan Silva and Rui Hou and Rui Wang and Saghar Hosseini and Sahana Chennabasappa and Sanjay Singh and Sean Bell and Seohyun Sonia Kim and Sergey Edunov and Shaoliang Nie and Sharan Narang and Sharath Raparthy and Sheng Shen and Shengye Wan and Shruti Bhosale and Shun Zhang and Simon Vandenhende and Soumya Batra and Spencer Whitman and Sten Sootla and Stephane Collot and Suchin Gururangan and Sydney Borodinsky and Tamar Herman and Tara Fowler and Tarek Sheasha and Thomas Georgiou and Thomas Scialom and Tobias Speckbacher and Todor Mihaylov and Tong Xiao and Ujjwal Karn and Vedanuj Goswami and Vibhor Gupta and Vignesh Ramanathan and Viktor Kerkez and Vincent Gonguet and Virginie Do and Vish Vogeti and Vítor Albiero and Vladan Petrovic and Weiwei Chu and Wenhan Xiong and Wenyin Fu and Whitney Meers and Xavier Martinet and Xiaodong Wang and Xiaofang Wang and Xiaoqing Ellen Tan and Xide Xia and Xinfeng Xie and Xuchao Jia and Xuewei Wang and Yaelle Goldschlag and Yashesh Gaur and Yasmine Babaei and Yi Wen and Yiwen Song and Yuchen Zhang and Yue Li and Yuning Mao and Zacharie Delpierre Coudert and Zheng Yan and Zhengxing Chen and Zoe Papakipos and Aaditya Singh and Aayushi Srivastava and Abha Jain and Adam Kelsey and Adam Shajnfeld and Adithya Gangidi and Adolfo Victoria and Ahuva Goldstand and Ajay Menon and Ajay Sharma and Alex Boesenberg and Alexei Baevski and Allie Feinstein and Amanda Kallet and Amit Sangani and Amos Teo and Anam Yunus and Andrei Lupu and Andres Alvarado and Andrew Caples and Andrew Gu and Andrew Ho and Andrew Poulton and Andrew Ryan and Ankit Ramchandani and Annie Dong and Annie Franco and Anuj Goyal and Aparajita Saraf and Arkabandhu Chowdhury and Ashley Gabriel and Ashwin Bharambe and Assaf Eisenman and Azadeh Yazdan and Beau James and Ben Maurer and Benjamin Leonhardi and Bernie Huang and Beth Loyd and Beto De Paola and Bhargavi Paranjape and Bing Liu and Bo Wu and Boyu Ni and Braden Hancock and Bram Wasti and Brandon Spence and Brani Stojkovic and Brian Gamido and Britt Montalvo and Carl Parker and Carly Burton and Catalina Mejia and Ce Liu and Changhan Wang and Changkyu Kim and Chao Zhou and Chester Hu and Ching-Hsiang Chu and Chris Cai and Chris Tindal and Christoph Feichtenhofer and Cynthia Gao and Damon Civin and Dana Beaty and Daniel Kreymer and Daniel Li and David Adkins and David Xu and Davide Testuggine and Delia David and Devi Parikh and Diana Liskovich and Didem Foss and Dingkang Wang and Duc Le and Dustin Holland and Edward Dowling and Eissa Jamil and Elaine Montgomery and Eleonora Presani and Emily Hahn and Emily Wood and Eric-Tuan Le and Erik Brinkman and Esteban Arcaute and Evan Dunbar and Evan Smothers and Fei Sun and Felix Kreuk and Feng Tian and Filippos Kokkinos and Firat Ozgenel and Francesco Caggioni and Frank Kanayet and Frank Seide and Gabriela Medina Florez and Gabriella Schwarz and Gada Badeer and Georgia Swee and Gil Halpern and Grant Herman and Grigory Sizov and Guangyi and Zhang and Guna Lakshminarayanan and Hakan Inan and Hamid Shojanazeri and Han Zou and Hannah Wang and Hanwen Zha and Haroun Habeeb and Harrison Rudolph and Helen Suk and Henry Aspegren and Hunter Goldman and Hongyuan Zhan and Ibrahim Damlaj and Igor Molybog and Igor Tufanov and Ilias Leontiadis and Irina-Elena Veliche and Itai Gat and Jake Weissman and James Geboski and James Kohli and Janice Lam and Japhet Asher and Jean-Baptiste Gaya and Jeff Marcus and Jeff Tang and Jennifer Chan and Jenny Zhen and Jeremy Reizenstein and Jeremy Teboul and Jessica Zhong and Jian Jin and Jingyi Yang and Joe Cummings and Jon Carvill and Jon Shepard and Jonathan McPhie and Jonathan Torres and Josh Ginsburg and Junjie Wang and Kai Wu and Kam Hou U and Karan Saxena and Kartikay Khandelwal and Katayoun Zand and Kathy Matosich and Kaushik Veeraraghavan and Kelly Michelena and Keqian Li and Kiran Jagadeesh and Kun Huang and Kunal Chawla and Kyle Huang and Lailin Chen and Lakshya Garg and Lavender A and Leandro Silva and Lee Bell and Lei Zhang and Liangpeng Guo and Licheng Yu and Liron Moshkovich and Luca Wehrstedt and Madian Khabsa and Manav Avalani and Manish Bhatt and Martynas Mankus and Matan Hasson and Matthew Lennie and Matthias Reso and Maxim Groshev and Maxim Naumov and Maya Lathi and Meghan Keneally and Miao Liu and Michael L. Seltzer and Michal Valko and Michelle Restrepo and Mihir Patel and Mik Vyatskov and Mikayel Samvelyan and Mike Clark and Mike Macey and Mike Wang and Miquel Jubert Hermoso and Mo Metanat and Mohammad Rastegari and Munish Bansal and Nandhini Santhanam and Natascha Parks and Natasha White and Navyata Bawa and Nayan Singhal and Nick Egebo and Nicolas Usunier and Nikhil Mehta and Nikolay Pavlovich Laptev and Ning Dong and Norman Cheng and Oleg Chernoguz and Olivia Hart and Omkar Salpekar and Ozlem Kalinli and Parkin Kent and Parth Parekh and Paul Saab and Pavan Balaji and Pedro Rittner and Philip Bontrager and Pierre Roux and Piotr Dollar and Polina Zvyagina and Prashant Ratanchandani and Pritish Yuvraj and Qian Liang and Rachad Alao and Rachel Rodriguez and Rafi Ayub and Raghotham Murthy and Raghu Nayani and Rahul Mitra and Rangaprabhu Parthasarathy and Raymond Li and Rebekkah Hogan and Robin Battey and Rocky Wang and Russ Howes and Ruty Rinott and Sachin Mehta and Sachin Siby and Sai Jayesh Bondu and Samyak Datta and Sara Chugh and Sara Hunt and Sargun Dhillon and Sasha Sidorov and Satadru Pan and Saurabh Mahajan and Saurabh Verma and Seiji Yamamoto and Sharadh Ramaswamy and Shaun Lindsay and Shaun Lindsay and Sheng Feng and Shenghao Lin and Shengxin Cindy Zha and Shishir Patil and Shiva Shankar and Shuqiang Zhang and Shuqiang Zhang and Sinong Wang and Sneha Agarwal and Soji Sajuyigbe and Soumith Chintala and Stephanie Max and Stephen Chen and Steve Kehoe and Steve Satterfield and Sudarshan Govindaprasad and Sumit Gupta and Summer Deng and Sungmin Cho and Sunny Virk and Suraj Subramanian and Sy Choudhury and Sydney Goldman and Tal Remez and Tamar Glaser and Tamara Best and Thilo Koehler and Thomas Robinson and Tianhe Li and Tianjun Zhang and Tim Matthews and Timothy Chou and Tzook Shaked and Varun Vontimitta and Victoria Ajayi and Victoria Montanez and Vijai Mohan and Vinay Satish Kumar and Vishal Mangla and Vlad Ionescu and Vlad Poenaru and Vlad Tiberiu Mihailescu and Vladimir Ivanov and Wei Li and Wenchen Wang and Wenwen Jiang and Wes Bouaziz and Will Constable and Xiaocheng Tang and Xiaojian Wu and Xiaolan Wang and Xilun Wu and Xinbo Gao and Yaniv Kleinman and Yanjun Chen and Ye Hu and Ye Jia and Ye Qi and Yenda Li and Yilin Zhang and Ying Zhang and Yossi Adi and Youngjin Nam and Yu and Wang and Yu Zhao and Yuchen Hao and Yundi Qian and Yunlu Li and Yuzi He and Zach Rait and Zachary DeVito and Zef Rosnbrick and Zhaoduo Wen and Zhenyu Yang and Zhiwei Zhao and Zhiyu Ma},
      year={2024},
      eprint={2407.21783},
      archivePrefix={arXiv},
      primaryClass={cs.AI},
      url={https://arxiv.org/abs/2407.21783}, 
}

@misc{gemma2025gemma3,
      title={{Gemma 3} Technical Report},
      author={{Gemma Team} and Aishwarya Kamath and Johan Ferret and Shreya Pathak and Nino Vieillard and Ramona Merhej and Sarah Perrin and Tatiana Matejovicova and Alexandre Ramé and Morgane Rivière and Louis Rouillard and Thomas Mesnard and Geoffrey Cideron and Jean-bastien Grill and Sabela Ramos and Edouard Yvinec and Michelle Casbon and Etienne Pot and Ivo Penchev and Gaël Liu and Francesco Visin and Kathleen Kenealy and Lucas Beyer and Xiaohai Zhai and Anton Tsitsulin and Robert Busa-Fekete and Alex Feng and Noveen Sachdeva and Benjamin Coleman and Yi Gao and Basil Mustafa and Iain Barr and Emilio Parisotto and David Tian and Matan Eyal and Colin Cherry and Jan-Thorsten Peter and Danila Sinopalnikov and Surya Bhupatiraju and Rishabh Agarwal and Mehran Kazemi and Dan Malkin and Ravin Kumar and David Vilar and Idan Brusilovsky and Jiaming Luo and Andreas Steiner and Abe Friesen and Abhanshu Sharma and Abheesht Sharma and Adi Mayrav Gilady and Adrian Goedeckemeyer and Alaa Saade and Alex Feng and Alexander Kolesnikov and Alexei Bendebury and Alvin Abdagic and Amit Vadi and András György and André Susano Pinto and Anil Das and Ankur Bapna and Antoine Miech and Antoine Yang and Antonia Paterson and Ashish Shenoy and Ayan Chakrabarti and Bilal Piot and Bo Wu and Bobak Shahriari and Bryce Petrini and Charlie Chen and Charline Le Lan and Christopher A. Choquette-Choo and CJ Carey and Cormac Brick and Daniel Deutsch and Danielle Eisenbud and Dee Cattle and Derek Cheng and Dimitris Paparas and Divyashree Shivakumar Sreepathihalli and Doug Reid and Dustin Tran and Dustin Zelle and Eric Noland and Erwin Huizenga and Eugene Kharitonov and Frederick Liu and Gagik Amirkhanyan and Glenn Cameron and Hadi Hashemi and Hanna Klimczak-Plucińska and Harman Singh and Harsh Mehta and Harshal Tushar Lehri and Hussein Hazimeh and Ian Ballantyne and Idan Szpektor and Ivan Nardini and Jean Pouget-Abadie and Jetha Chan and Joe Stanton and John Wieting and Jonathan Lai and Jordi Orbay and Joseph Fernandez and Josh Newlan and Ju-yeong Ji and Jyotinder Singh and Kat Black and Kathy Yu and Kevin Hui and Kiran Vodrahalli and Klaus Greff and Linhai Qiu and Marcella Valentine and Marina Coelho and Marvin Ritter and Matt Hoffman and Matthew Watson and Mayank Chaturvedi and Michael Moynihan and Min Ma and Nabila Babar and Natasha Noy and Nathan Byrd and Nick Roy and Nikola Momchev and Nilay Chauhan and Noveen Sachdeva and Oskar Bunyan and Pankil Botarda and Paul Caron and Paul Kishan Rubenstein and Phil Culliton and Philipp Schmid and Pier Giuseppe Sessa and Pingmei Xu and Piotr Stanczyk and Pouya Tafti and Rakesh Shivanna and Renjie Wu and Renke Pan and Reza Rokni and Rob Willoughby and Rohith Vallu and Ryan Mullins and Sammy Jerome and Sara Smoot and Sertan Girgin and Shariq Iqbal and Shashir Reddy and Shruti Sheth and Siim Põder and Sijal Bhatnagar and Sindhu Raghuram Panyam and Sivan Eiger and Susan Zhang and Tianqi Liu and Trevor Yacovone and Tyler Liechty and Uday Kalra and Utku Evci and Vedant Misra and Vincent Roseberry and Vlad Feinberg and Vlad Kolesnikov and Woohyun Han and Woosuk Kwon and Xi Chen and Yinlam Chow and Yuvein Zhu and Zichuan Wei and Zoltan Egyed and Victor Cotruta and Minh Giang and Phoebe Kirk and Anand Rao and Kat Black and Nabila Babar and Jessica Lo and Erica Moreira and Luiz Gustavo Martins and Omar Sanseviero and Lucas Gonzalez and Zach Gleicher and Tris Warkentin and Vahab Mirrokni and Evan Senter and Eli Collins and Joelle Barral and Zoubin Ghahramani and Raia Hadsell and Yossi Matias and D. Sculley and Slav Petrov and Noah Fiedel and Noam Shazeer and Oriol Vinyals and Jeff Dean and Demis Hassabis and Koray Kavukcuoglu and Clement Farabet and Elena Buchatskaya and Jean-Baptiste Alayrac and Rohan Anil and Dmitry and Lepikhin and Sebastian Borgeaud and Olivier Bachem and Armand Joulin and Alek Andreev and Cassidy Hardin and Robert Dadashi and Léonard Hussenot},
      year={2025},
      eprint={2503.19786},
      archivePrefix={arXiv},
      primaryClass={cs.CL},
      url={https://arxiv.org/abs/2503.19786}, 
}
\clearpage
\appendix

\section{Proofs}
\label{app:proofs}

This appendix provides the proofs and estimation error analysis for DWE.
It also derives the reverse-KL logit gradient used in the OPD experiments.

Throughout this section, the context $s$, student parameters $\theta$,
and teacher distribution are fixed. Expectations are over sampling randomness.

\subsection{Zero-Sum Property}
\label{app:zero-sum}

We prove Lemma~\ref{lem:zero-sum}, which ensures that the positive and
negative parts of the logit gradient have equal total magnitude. This
property allows DWE to use a common weight for each sampled pair.

\begin{proof}
Let $p=\operatorname{softmax}(z)$. Its derivatives satisfy
$\partial p_j/\partial z_i=p_j(\delta_{ij}-p_i)$, where $\delta_{ij}$
is the Kronecker delta. Since $q$ is fixed and the loss depends on $z$
through $p$, the chain rule gives
\begin{equation}
    \begin{aligned}
    g_i
        &=\sum_j\frac{\partial\ell}{\partial p_j}
            p_j(\delta_{ij}-p_i)\\
        &=p_i\frac{\partial\ell}{\partial p_i}
          -p_i\sum_j p_j\frac{\partial\ell}{\partial p_j}.
    \end{aligned}
    \label{eq:app-softmax-gradient}
\end{equation}
Summing over all logit coordinates and using $\sum_i p_i=1$, we obtain
\begin{equation}
    \sum_i g_i
        =\sum_i p_i\frac{\partial\ell}{\partial p_i}
          -\left(\sum_i p_i\right)
           \sum_j p_j\frac{\partial\ell}{\partial p_j}
        =0.
    \label{eq:app-zero-sum}
\end{equation}
Thus $\sum_i(g_+)_i=\sum_i(g_-)_i$, giving the common magnitude
$m=\tfrac12\|g\|_1$ used in the paired construction.
\end{proof}

\subsection{Unbiasedness}
\label{app:stratified-unbiased}

We show that the DWE estimator preserves the exact logit gradient in
expectation after shared-quantile pairing and stratified sampling. We then
extend this result to parameter gradients on a fixed minibatch.

\begin{theorem}[Unbiasedness]
\label{thm:dwe-estimator}
Conditional on the context $s$ and current parameters $\theta$, the
estimator in Eq.~\eqref{eq:dwe-estimator} satisfies
\begin{equation}
    \mathbb E[\widehat g\mid s,\theta]=g.
    \label{eq:dwe-unbiased}
\end{equation}
\end{theorem}

\begin{proof}
When $g=0$, DWE uses the zero estimator and the result holds directly.
For $g\neq0$, we first verify that shared-quantile sampling preserves the
two marginal distributions. Fix a vocabulary order $\tau$. The generalized
inverse CDF
$F^{-1}(u)=\min\{j:F(j)\geq u\}$ returns a position in the ordered
vocabulary. If token $i$ has ordered position $j$, it is selected by
$I(U)=\tau(F_{r_+,\tau}^{-1}(U))$ precisely when $U$ lies in the
corresponding CDF interval, whose length is $(r_+)_i$.
Hence $\Pr(I(U)=i)=(r_+)_i$ for $U\sim\operatorname{Unif}(0,1)$;
likewise, $\Pr(J(U)=i)=(r_-)_i$. Zero-probability entries have empty
intervals, and the choice of interval endpoints has probability zero.
The shared quantile therefore defines a coupling with the required
marginals. For $f(u)=m(e_{I(u)}-e_{J(u)})$,
\begin{equation}
    \int_0^1 f(u)\,\mathrm du
      =m\bigl(\mathbb E[e_{I(U)}]-\mathbb E[e_{J(U)}]\bigr)
      =m(r_+-r_-)=g.
    \label{eq:app-quantile-mean}
\end{equation}

We next consider stratification. Let $A_k=((k-1)/K,k/K)$. Each interval
has length $1/K$, so a uniform sample $U_k$ within $A_k$ satisfies
$\mathbb E[f(U_k)]=K\int_{A_k}f(u)\,\mathrm du$.

Averaging across the $K$ strata yields
\begin{equation}
    \mathbb E[\widehat g]
       =\frac1K\sum_{k=1}^K\mathbb E[f(U_k)]
       =\sum_{k=1}^K\int_{A_k}f(u)\,\mathrm du=g.
    \label{eq:app-stratified-mean}
\end{equation}
Thus averaging over all strata recovers $g$ in expectation, even though
an individual stratum need not preserve the full marginals.
\end{proof}

\paragraph{Minibatch gradients.}
For fixed inputs and parameters, backpropagation is linear in the logit
gradients. Their unbiasedness therefore extends to the parameter gradient
and its weighted sum over a minibatch. Let $\mathcal B$ be a fixed minibatch
of training sequences,
and let $\omega$ denote DWE's sampling randomness. Write
$\widehat{\nabla_\theta\mathcal L_{\mathcal B}}$ for the parameter
gradient formed using the DWE estimator and $\mathcal L_{\mathcal B}$ for
the full-vocabulary distillation loss on this batch. The token-loss
normalization and minibatch weights are fixed given $\mathcal B$, so
Eq.~\eqref{eq:parameter-unbiased} and linearity give
\begin{equation}
    \mathbb E_{\omega}\!\left[
        \widehat{\nabla_\theta\mathcal L_{\mathcal B}}
        \mid \mathcal B,\theta
    \right]
    =\nabla_\theta\mathcal L_{\mathcal B}.
    \label{eq:minibatch-unbiased}
\end{equation}
Thus, on a fixed minibatch, DWE recovers the full-vocabulary parameter
gradient in expectation.

\subsection{Variance Reduction}
\label{app:stratified-variance}

We analyze variance reduction from paired and stratified sampling, then
derive the estimation error bound in Eq.~\eqref{eq:estimation-error-bound}.

\paragraph{Paired sampling.}
For $g\neq0$, IS draws $L_1,\ldots,L_{2K}$ independently with
replacement from $\rho_v=|g_v|/(2m)$ and forms
\begin{equation}
    \widehat g_{\mathrm{IS}}
    =\frac{m}{K}\sum_{a=1}^{2K}
        \operatorname{sgn}(g_{L_a})e_{L_a}.
    \label{eq:app-coordinate-is}
\end{equation}
Let $\widehat g_{\mathrm{ind}}$ instead average $K$ pairs with
$I_k\sim r_+$ and $J_k\sim r_-$, with all draws independent.
Both estimators use $2K$ coordinate draws and have expectation $g$.
Their covariance matrices are
\begin{equation}
    \begin{aligned}
    \operatorname{Cov}(\widehat g_{\mathrm{IS}})
        &=\frac{m}{K}\operatorname{diag}(|g|)
          -\frac{gg^\top}{2K},\\
    \operatorname{Cov}(\widehat g_{\mathrm{ind}})
        &=\frac{m}{K}\operatorname{diag}(|g|)
          -\frac{m^2}{K}
            \left(r_+r_+^\top+r_-r_-^\top\right).
    \end{aligned}
    \label{eq:app-sampling-covariances}
\end{equation}
Subtracting and using $g=m(r_+-r_-)$ and $|g|=m(r_++r_-)$ gives
\begin{equation}
    \operatorname{Cov}(\widehat g_{\mathrm{IS}})
    -\operatorname{Cov}(\widehat g_{\mathrm{ind}})
    =\frac{|g|\,|g|^\top}{2K}\succeq0.
    \label{eq:app-pairing-variance}
\end{equation}
This term is the covariance due to random allocation of IS draws between
the two signs. Sampling exactly $K$ coordinates from each part removes it.

\paragraph{Stratified sampling.}
For a fixed coupling and $K$ pairs, stratification removes the variance
caused by random allocation of samples across strata.

\begin{theorem}[Variance reduction]
\label{thm:variance-reduction}
For a fixed coupling and $K$ pairs, let
$\widehat g_{\mathrm{strat}}$ denote the estimator in
Eq.~\eqref{eq:dwe-estimator}, and let $\widehat g_{\mathrm{iid}}$ average
$K$ paired estimates obtained from independent uniform quantile draws. Then
\begin{equation}
    \operatorname{Cov}(\widehat g_{\mathrm{strat}}\mid s,\theta)
    \preceq
    \operatorname{Cov}(\widehat g_{\mathrm{iid}}\mid s,\theta).
    \label{eq:stratified-covariance}
\end{equation}
\end{theorem}
Here $\preceq$ denotes the positive-semidefinite order.

\begin{proof}
Let $X=f(U)$ for $U\sim\operatorname{Unif}(0,1)$, and let $S$ index the stratum
containing $U$. Each stratum has probability $1/K$. Define its conditional
mean and covariance as
\[
    \begin{aligned}
    \mu_k&=\mathbb E[X\mid S=k],\\
    \Sigma_k&=\operatorname{Cov}(X\mid S=k).
    \end{aligned}
\]
The law of total covariance separates variation within strata from
variation between their means:
\begin{equation}
    \operatorname{Cov}(X)
    =\frac1K\sum_{k=1}^K\Sigma_k+\operatorname{Cov}(\mu_S).
    \label{eq:app-total-covariance}
\end{equation}
For $K$ independent draws from the full coupling, averaging divides this
covariance by $K$. Stratification instead takes one independent draw
$X_k$ from the conditional distribution of $X$ given $S=k$ in each
stratum. The two estimators satisfy
\[
    \begin{aligned}
    \operatorname{Cov}(\widehat g_{\mathrm{iid}})
        &=\frac1{K^2}\sum_{k=1}^K\Sigma_k
          +\frac1K\operatorname{Cov}(\mu_S),\\
    \operatorname{Cov}(\widehat g_{\mathrm{strat}})
        &=\frac1{K^2}\sum_{k=1}^K\Sigma_k.
    \end{aligned}
\]
Subtracting gives the covariance improvement
\begin{equation}
    \operatorname{Cov}(\widehat g_{\mathrm{iid}})
    -\operatorname{Cov}(\widehat g_{\mathrm{strat}})
    =\frac1K\operatorname{Cov}(\mu_S)\succeq0.
    \label{eq:app-covariance-reduction}
\end{equation}
Thus stratification has no larger covariance than iid sampling from the
same coupling. It removes the component caused by random allocation of
samples across strata.
\end{proof}

\paragraph{Estimation error.}
For any fixed coupling, the positive and negative supports are disjoint,
so $\|f(U)\|_2^2=2m^2$. Since $\mathbb E[f(U)]=g$, averaging $K$
independent paired samples has mean-squared error
$(2m^2-\|g\|_2^2)/K$. Taking traces in
Theorem~\ref{thm:variance-reduction} therefore gives
\begin{equation}
    \begin{aligned}
    \mathbb E\!\left[\|\widehat g-g\|_2^2\right]
        &\leq\frac{2m^2-\|g\|_2^2}{K}\\
        &=\mathbb E\!\left[\|\widehat g_{\mathrm{IS}}-g\|_2^2\right]
          -\frac{\|g\|_2^2}{2K}.
    \end{aligned}
    \label{eq:app-error-comparison}
\end{equation}
The equality follows from the IS covariance in
Eq.~\eqref{eq:app-sampling-covariances}. For $g=0$, the zero estimator
satisfies the error bound directly.

\subsection{KL Gradients}
\label{app:kl-gradients}

The main text uses forward KL. Here we derive the reverse-KL logit gradient
used in the OPD experiments.

\paragraph{Reverse KL.}
\label{app:rkl}
For $\ell=D_{\mathrm{KL}}(p\|q)=\sum_j p_j\log(p_j/q_j)$,
we have $\partial\ell/\partial p_j=\log(p_j/q_j)+1$.
Substituting into Eq.~\eqref{eq:app-softmax-gradient} gives
\begin{equation}
    \begin{aligned}
    g_i
      &=p_i\left(\log\frac{p_i}{q_i}+1\right)
          -p_i\sum_j p_j\left(\log\frac{p_j}{q_j}+1\right)\\
      &=p_i\left(\log\frac{p_i}{q_i}-\ell\right).
    \end{aligned}
    \label{eq:app-reverse-kl}
\end{equation}
Since $\sum_i g_i=0$, the construction in Section~\ref{sec:method}
also applies to reverse KL.

\clearpage
\section{Experimental Details and Additional Results}
\label{app:experiment-details}

\subsection{Experimental Configuration}
\label{app:experimental-configuration}

\subsubsection{Training}
\label{app:training-configuration}

For off-policy KD, we sample 80M tokens from the
\texttt{sample-10BT} subset of FineWeb-Edu~\citep{lozhkov2024finewebedu},
following the training budget used in SE-KD~\citep{tavor2026rethinking}.
Table~\ref{tab:kd-training-config} summarizes the training configuration
shared across student sizes.

\begin{table}[ht]
  \centering
  \caption{\textbf{KD training configuration.} Shared settings across student sizes.}
  \label{tab:kd-training-config}
  \vspace{5pt}
  \small
  \renewcommand{\arraystretch}{1.15}
  \begin{tabular*}{\textwidth}{@{\extracolsep{\fill}}llll@{}}
    \toprule
    \multicolumn{2}{l}{\textbf{Optimization}} &
    \multicolumn{2}{l}{\textbf{Implementation and methods}} \\
    \cmidrule(lr){1-2}\cmidrule(lr){3-4}
    Training tokens & 80M & Hardware & $1\times$ H100 80GB \\
    Data sampling seed & 2026 & Sequence length & 4,096 \\
    Training seeds & 42, 43, 44 & & \\
    Microbatch size & 2 & Projection precision & BF16 \\
    Gradient accumulation & 4 & Probability/residual precision & FP32 \\
    Optimizer & 8-bit Adam & Token chunk size & 256 \\
    Learning rate & $10^{-5}$ & Gradient checkpointing & Enabled \\
    Learning-rate schedule & Constant & Distillation loss & Forward KL \\
    Warmup updates & 100 & Temperature & 1 \\
    Weight decay & 0 & DWE pairs $K$ & 8 \\
    Gradient clipping norm & 1 & IS coordinate draws & 16 \\
    \bottomrule
  \end{tabular*}
\end{table}

\FloatBarrier

For OPD, we use Qwen3-1.7B~\citep{yang2025qwen3} as the student and
Qwen3-4B-Instruct-2507~\citep{yang2025qwen3} as the teacher, with thinking
disabled during training and evaluation. Following OPD$^2$~\citep{heo2026opd2},
we construct a pool of approximately 100K prompts sampled equally from
OpenMathReasoning~\citep{moshkov2025aimo2},
OpenCodeReasoning~\citep{ahmad2025opencode}, and
OpenScienceReasoning-2~\citep{nvidia2025science}. We use only the prompts
and generate responses on-policy. We compare \method{} with full-vocabulary
reverse KL (Full).
Table~\ref{tab:opd-training-config} lists the OPD settings.
Within each setting, all methods share the student initialization and
numerical settings. Full reduces memory use by recomputing output-head
activations during backpropagation.

\begin{table}[ht]
  \centering
  \caption{\textbf{OPD training configuration.} Shared settings and method-specific sampling parameters.}
  \label{tab:opd-training-config}
  \vspace{5pt}
  \small
  \renewcommand{\arraystretch}{1.15}
  \begin{tabular*}{\textwidth}{@{\extracolsep{\fill}}llll@{}}
    \toprule
    \multicolumn{2}{l}{\textbf{Optimization and generation}} &
    \multicolumn{2}{l}{\textbf{Implementation and methods}} \\
    \cmidrule(lr){1-2}\cmidrule(lr){3-4}
    Updates & 100 & Hardware & $1\times$ H100 80GB \\
    Training seeds & 42, 43, 44 & Projection precision & BF16 \\
    Global batch size & 256 & Probability/residual precision & FP32 \\
    Microbatch size & 2 & Token chunk size & 256 \\
    Gradient accumulation & 128 & Gradient checkpointing & Enabled \\
    Peak learning rate & $5\times10^{-6}$ & Dropout & Disabled \\
    Learning-rate schedule & Cosine decay & DeepSpeed & ZeRO-2 \\
    Final learning rate & $5\times10^{-7}$ & Generation engine & vLLM, colocated \\
    Warmup updates & 10 & Training precision & BF16 \\
    Sampling temperature & 0.7 & DWE sampling & $K=8$, geometry-aware order \\
    Maximum response tokens & 8,192 & & \\
    Thinking & Disabled & & \\
    \bottomrule
  \end{tabular*}
\end{table}

\clearpage
\subsubsection{Evaluation}
\label{app:evaluation-configuration}

KD evaluation uses LM Evaluation Harness v0.4.13~\citep{eval-harness},
BF16 computation, and no chat template. Table~\ref{tab:kd-benchmark-config}
lists the complete zero-shot evaluation sets and scoring metrics.
GSM8K uses greedy decoding with at most 512 new tokens and exact-match
scoring with flexible answer extraction.

\begin{table}[ht]
  \centering
  \caption{\textbf{KD evaluation benchmarks.} Question counts refer to
  the complete evaluated sets.}
  \label{tab:kd-benchmark-config}
  \vspace{5pt}
  \small
  \renewcommand{\arraystretch}{1.12}
  \begin{tabular*}{.94\textwidth}{@{\extracolsep{\fill}}lrcl@{}}
    \toprule
    Benchmark & Questions & Repeats & Metric \\
    \midrule
    HellaSwag~\citep{zellers2019hellaswag} & 10,042 & 1 & Length-normalized accuracy \\
    PIQA~\citep{bisk2020piqa} & 1,838 & 1 & Accuracy \\
    ARC-Easy~\citep{clark2018arc} & 2,376 & 1 & Accuracy \\
    ARC-Challenge~\citep{clark2018arc} & 1,172 & 1 & Length-normalized accuracy \\
    GSM8K~\citep{cobbe2021verifiers} & 1,319 & 1 & Exact match \\
    LAMBADA~\citep{paperno2016lambada} & 5,153 & 1 & Accuracy \\
    MMLU~\citep{hendrycks2021mmlu} & 14,042 & 1 & Accuracy \\
    \bottomrule
  \end{tabular*}
\end{table}

For OPD, we use the evaluation code from OPD$^2$~\citep{heo2026opd2}
and Evalchemy~\citep{guha2024evalchemy}. We use BF16 inference with a
maximum generation length of 32,768 tokens.
The sampling temperature is 0.2 for CodeContests and 0.7 otherwise.
Table~\ref{tab:opd-benchmark-config} lists the evaluated sets;
LiveCodeBench v5 uses \texttt{LCBv5-v2}. For SuperGPQA, we sample 50
questions without replacement per discipline using seed 0, giving the same 650-question
subset for all methods.

\begin{table}[ht]
  \centering
  \caption{\textbf{OPD evaluation benchmarks.} Question counts refer to the evaluated sets.}
  \label{tab:opd-benchmark-config}
  \vspace{5pt}
  \small
  \setlength{\tabcolsep}{1pt}
  \renewcommand{\arraystretch}{1.12}
  \begin{tabular*}{\textwidth}{@{\extracolsep{\fill}}llrcl@{}}
    \toprule
    Domain & Benchmark & Questions & Repeats & Metric \\
    \midrule
    Mathematics & MATH500~\citep{hendrycks2021math,lightman2023verify} & 500 & 10 & Accuracy \\
    & AIME25~\citep{maa2025aime} & 30 & 16 & Accuracy \\
    & AMC23~\citep{maa2023amc} & 40 & 16 & Accuracy \\
    & HMMT25~\citep{matharena2025hmmt} & 30 & 16 & Accuracy \\
    & OlympiadBench~\citep{he2024olympiad} & 674 & 3 & Accuracy \\
    & RG-Math~\citep{stojanovski2025gym} & 100 & 3 & Native task score \\
    \midrule
    Code & CodeContests~\citep{li2022alphacode} & 165 & 3 & Accuracy \\
    & Codeforces~\citep{openr12025codeforces} & 453 & 3 & Accuracy \\
    & LiveCodeBench v5~\citep{jain2024livecode} & 369 & 3 & Accuracy \\
    & RG-Algorithm~\citep{stojanovski2025gym} & 100 & 3 & Native task score \\
    \midrule
    Science & GPQA Diamond~\citep{rein2023gpqa} & 198 & 10 & Accuracy \\
    & SuperGPQA~\citep{team2025supergpqa} & 650 & 3 & Accuracy \\
    & SciBench~\citep{wang2023scibench} & 692 & 3 & Accuracy \\
    \bottomrule
  \end{tabular*}
\end{table}

\FloatBarrier
\begingroup
\setlength{\textfloatsep}{12pt plus 2pt minus 2pt}
\setlength{\intextsep}{6pt plus 2pt minus 2pt}
\setlength{\parskip}{3pt}
\clearpage
\subsection{Supplementary Analyses}
\label{app:supplementary-analyses}

We analyze off-policy KD training costs and report held-out evaluation results.

\subsubsection{Runtime Analysis}
\label{app:runtime-analysis}

We measure complete training updates for Qwen3-1.7B~\citep{yang2025qwen3}
under the configuration in Section~\ref{sec:experimental-setup}.
Results average ten updates after three warmup updates.

\begin{table}[!ht]
  \centering
  \caption{\textbf{KD update runtime.} We report the mean runtime of each
  stage of a complete training update in seconds. The forward pass includes
  loss computation and estimator preparation. Optimizer time includes gradient
  clipping and parameter updates, and other overhead includes data loading.}
  \label{tab:runtime-analysis}
  \vspace{5pt}
  \small
  \renewcommand{\arraystretch}{1.12}
  \setlength{\tabcolsep}{7pt}
  \begin{tabular}{@{}lrrrrr@{}}
    \toprule
    Method & Forward & Backward & Optimizer & Other & Total \\
    \midrule
Full & 0.648 & 1.522 & 0.069 & 0.018 & 2.258 \\
DWE & 0.738 & 1.079 & 0.067 & 0.017 & 1.901 \\
    \bottomrule
  \end{tabular}
\end{table}

Table~\ref{tab:runtime-analysis} shows the time breakdown of a complete
training update. DWE reduces backward time from 1.522\,s to 1.079\,s
(29.1\%), while forward time increases from 0.648\,s to 0.738\,s for
estimator preparation.
The backward savings outweigh this overhead, reducing update time from
2.258\,s to 1.901\,s, a 15.8\% reduction. Over the full 80M-token training
runs, training time decreases by 15.7\%.

\FloatBarrier
\subsubsection{Held-out Evaluation}
\label{app:kd-heldout}

We evaluate teacher matching and next-token prediction on held-out test
text using exact full-vocabulary forward KL and perplexity, respectively.

\begin{table}[!ht]
  \centering
  \caption{\textbf{Held-out evaluation.} Final seed 42 checkpoints.}
  \label{tab:kd-heldout}
  \vspace{5pt}
  \small
  \renewcommand{\arraystretch}{1.12}
  \setlength{\tabcolsep}{5pt}
  \begin{tabular}{@{}lrr@{\hspace{1.5em}}rr@{\hspace{1.5em}}rr@{}}
    \toprule
    & \multicolumn{2}{c}{{\footnotesize\color{inkmuted}4B-Instruct $\to$ 0.6B}}
    & \multicolumn{2}{c}{{\footnotesize\color{inkmuted}8B $\to$ 1.7B}}
    & \multicolumn{2}{c}{{\footnotesize\color{inkmuted}32B $\to$ 4B}} \\
    \cmidrule(lr){2-3}\cmidrule(lr){4-5}\cmidrule(lr){6-7}
    Method & Forward KL & Perplexity & Forward KL & Perplexity
      & Forward KL & Perplexity \\
    \midrule
Full & 0.5207 & 17.238 & 0.3661 & 12.929 & 0.3947 & 10.505 \\
IS & 0.5223 & 17.254 & 0.3668 & 12.935 & 0.3957 & 10.518 \\
DWE & 0.5213 & 17.242 & 0.3663 & 12.928 & 0.3950 & 10.508 \\
    \bottomrule
  \end{tabular}
\end{table}

\FloatBarrier
\endgroup
\begingroup
\setlength{\parskip}{4pt}
\setlength{\textfloatsep}{12pt plus 2pt minus 2pt}
\setlength{\intextsep}{10pt plus 2pt minus 2pt}
\subsection{On-Policy Distillation}
\label{app:opd-experiments}
\newcommand{\OPDTrainingReductionMin}{1.21}
\newcommand{\OPDTrainingReductionMax}{2.12}
\newcommand{\OPDProfileBwdSaved}{17.17}
\newcommand{\OPDProfileForwardAdded}{1.94}
\newcommand{\OPDProfileRolloutTeacherShare}{53.2}

We additionally evaluate \method{} in OPD, where training contexts are
generated by the student. The configuration is given in
Appendix~\ref{app:training-configuration}.

\subsubsection{Output-Head Efficiency}
\label{app:opd-head-efficiency}

\begingroup
\begin{wrapfigure}{r}{.50\textwidth}
  \vspace{-19pt}
  \centering
  \setlength{\abovecaptionskip}{4pt}
  \includegraphics[width=.90\linewidth]{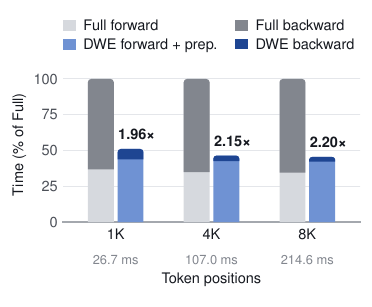}\par
  \caption{\textbf{Output-head time.} Runtimes are normalized to Full.
  Labels above the bars give total speedups, and times below give Full's
  absolute runtime.}
  \label{fig:opd-head-efficiency}
\end{wrapfigure}

We evaluate output-head savings in OPD on responses to 16 fresh prompts
per domain, generated by the final seed 42 DWE checkpoint with at most
8,192 tokens. We sample 8,192 distinct positions per domain; the 1K and 4K
settings use nested subsets. Figure~\ref{fig:opd-head-efficiency} averages
the three domains equally, following the timing protocol in
Section~\ref{sec:kd-head-efficiency}. The one-time geometry ordering is
excluded.

\method{} achieves a \OPDHeadSpeedMin--\OPDHeadSpeedMax$\times$ speedup in total
output-head computation, driven by an 8.6--20.2$\times$ speedup in the
backward pass. At 8K token positions, even with \OPDHeadFwdAddEightK\,ms of
additional forward computation for estimator preparation, the backward pass
saves \OPDHeadBwdSavedEightK\,ms, yielding a 54.5\% reduction in total time,
from \OPDHeadExactEightK\,ms to \OPDHeadDweEightK\,ms.

\par\WFclear
\endgroup

\FloatBarrier
\begingroup
\setlength{\intextsep}{4pt plus 1pt minus 1pt}
\subsubsection{Training Efficiency}
\label{app:opd-training-efficiency}

We evaluate end-to-end OPD training efficiency across three seeds.

\begin{table}[!ht]
  \centering
  \caption{\textbf{OPD training runtime.} We report mean seconds per update
  across seeds (top) and by stage (bottom). Stage timings average three
  updates after two warmups.}
  \label{tab:opd-runtime-analysis}
  \vspace{5pt}
  \small
  \setlength{\tabcolsep}{0pt}
  \renewcommand{\arraystretch}{1.12}
  \begin{tabular}{@{}w{l}{.22\textwidth}*{4}{w{c}{.195\textwidth}}@{}}
    \toprule
    \multicolumn{5}{@{}l}{\small\itshape\color{inkmuted}End-to-end training} \\[1pt]
    Method & Seed 42 & Seed 43 & Seed 44 & Mean \\
    \midrule
Full & 694.77 & 683.65 & 687.21 & 688.54 \\
\rowcolor{dwetint} DWE & 680.03 & 675.35 & 678.72 & 678.03 \\
Reduction (\%) & 2.12 & 1.21 & 1.24 & 1.53 \\
  \end{tabular}
  \par\vspace{1pt}
  {\color{rulegray}\hrule height .4pt}
  \vspace{1pt}
  \begin{tabular}{@{}w{l}{.16\textwidth}*{6}{w{c}{.14\textwidth}}@{}}
    \multicolumn{7}{@{}l}{\small\itshape\color{inkmuted}Update breakdown} \\[1pt]
    Method & Generation & \shortstack{Teacher\\backbone} & Forward & Backward &
      \shortstack{Optimizer\\and other} & Total \\
    \midrule
Full & 216.82 & 154.06 & 75.48 & 249.79 & 0.60 & 696.75 \\
\rowcolor{dwetint} DWE & 220.81 & 153.95 & 77.42 & 232.62 & 0.58 & 685.39 \\
    \bottomrule
  \end{tabular}
\end{table}

Table~\ref{tab:opd-runtime-analysis} shows that \method{} reduces update
time across all three seeds by
\OPDTrainingReductionMin--\OPDTrainingReductionMax\%, with a mean reduction
of \OPDStepSavedPct\%. In the stage profile, the backward pass saves
\OPDProfileBwdSaved\,s, exceeding the \OPDProfileForwardAdded\,s increase in
forward time, which includes estimator preparation. Response generation
and teacher-backbone computation account for
\OPDProfileRolloutTeacherShare\% of Full's update time, explaining why
end-to-end gains are smaller than output-head speedups.

\par
\FloatBarrier
\subsubsection{Reasoning Performance}
\label{app:opd-performance}

We evaluate reasoning performance on 13 benchmarks covering mathematics,
code, and science. Table~\ref{tab:opd-main-results} reports final-checkpoint
scores, with the initial student and teacher as references.

\method{} achieves an average score of \OPDOverallDwe{}, compared with
\OPDOverallIS{} for IS and \OPDOverallExact{} for Full. These results show
that \method{} maintains competitive downstream performance while using
sparse output-head backpropagation.

\vspace{-8pt}
\begin{table}[!ht]
\centering
\caption{\textbf{Reasoning performance across 13 benchmarks.} Scores (\%) are means with standard deviations over repeated evaluations. Superscripts give evaluation counts.
Best scores among Full, IS, and DWE are bolded. Average is the
unweighted mean across all 13 benchmarks.}
\label{tab:opd-main-results}
\vspace{5pt}
\setlength{\tabcolsep}{0pt}
\renewcommand{\arraystretch}{.8}
\setlength{\fboxsep}{0pt}
\newcommand{\opdbest}[1]{\rule[-3pt]{0pt}{12pt}\textbf{#1}}
\begin{NiceTabular}{>{\rule[-3pt]{0pt}{12pt}}w{l}{.25\textwidth}
    >{\color{inkmuted}}w{c}{.15\textwidth}
    >{\color{inkmuted}}w{c}{.15\textwidth}
    w{c}{.15\textwidth}
    w{c}{.15\textwidth}
    w{c}{.15\textwidth}}
  \rule{0pt}{14pt}\Block[l]{2-1}{\textbf{Benchmark}}
    & \Block{1-2}{\textbf{Reference models}} &
    & \Block{1-2}{\textbf{Baselines}} &
    & \Block{2-1}{\textbf{\method{}}} \\[1pt]
    & \Block{}{Initial}
    & \Block{}{Teacher}
    & \Block{}{Full}
    & \Block{}{IS}
    & \\[1pt]
\Block[l]{1-6}{\grouplabel{Mathematics}}\rule{0pt}{12.5pt} & & & & & \\[-2.5pt]
MATH500 & 69.6 & 90.2 & 82.32{\scriptsize\color{inkmuted}\,$\pm$0.77\textsuperscript{(10)}} & 82.34{\scriptsize\color{inkmuted}\,$\pm$0.78\textsuperscript{(10)}} & \opdbest{82.42}{\scriptsize\color{inkmuted}\,$\pm$1.34\textsuperscript{(10)}} \\
AIME25 & 0.0 & 36.7 & \opdbest{23.33}{\scriptsize\color{inkmuted}\,$\pm$3.85\textsuperscript{(16)}} & 22.71{\scriptsize\color{inkmuted}\,$\pm$4.25\textsuperscript{(16)}} & 22.71{\scriptsize\color{inkmuted}\,$\pm$3.27\textsuperscript{(16)}} \\
AMC23 & 40.0 & 92.5 & 71.56{\scriptsize\color{inkmuted}\,$\pm$5.15\textsuperscript{(16)}} & \opdbest{73.59}{\scriptsize\color{inkmuted}\,$\pm$5.08\textsuperscript{(16)}} & 73.44{\scriptsize\color{inkmuted}\,$\pm$6.05\textsuperscript{(16)}} \\
HMMT25 & 3.3 & 36.7 & \opdbest{15.83}{\scriptsize\color{inkmuted}\,$\pm$4.30\textsuperscript{(16)}} & \opdbest{15.83}{\scriptsize\color{inkmuted}\,$\pm$4.30\textsuperscript{(16)}} & \opdbest{15.83}{\scriptsize\color{inkmuted}\,$\pm$4.30\textsuperscript{(16)}} \\
OlympiadBench & 24.3 & 44.2 & \opdbest{36.55}{\scriptsize\color{inkmuted}\,$\pm$0.34\textsuperscript{(3)}} & 36.00{\scriptsize\color{inkmuted}\,$\pm$0.43\textsuperscript{(3)}} & \opdbest{36.55}{\scriptsize\color{inkmuted}\,$\pm$0.56\textsuperscript{(3)}} \\
RG-Math & 83.7 & 93.6 & 89.19{\scriptsize\color{inkmuted}\,$\pm$2.22\textsuperscript{(3)}} & \opdbest{90.63}{\scriptsize\color{inkmuted}\,$\pm$0.27\textsuperscript{(3)}} & 90.58{\scriptsize\color{inkmuted}\,$\pm$0.80\textsuperscript{(3)}} \\
\Block[l]{1-6}{\grouplabel{Code}}\rule{0pt}{12.5pt} & & & & & \\[-2.5pt]
CodeContests & 2.4 & 18.2 & 14.14{\scriptsize\color{inkmuted}\,$\pm$4.86\textsuperscript{(3)}} & \opdbest{14.95}{\scriptsize\color{inkmuted}\,$\pm$5.08\textsuperscript{(3)}} & 13.33{\scriptsize\color{inkmuted}\,$\pm$5.45\textsuperscript{(3)}} \\
Codeforces & 1.5 & 5.7 & \opdbest{15.23}{\scriptsize\color{inkmuted}\,$\pm$0.58\textsuperscript{(3)}} & 14.28{\scriptsize\color{inkmuted}\,$\pm$1.04\textsuperscript{(3)}} & 14.13{\scriptsize\color{inkmuted}\,$\pm$0.80\textsuperscript{(3)}} \\
LiveCodeBench v5 & 5.4 & 22.8 & 15.99{\scriptsize\color{inkmuted}\,$\pm$2.75\textsuperscript{(3)}} & 20.87{\scriptsize\color{inkmuted}\,$\pm$7.80\textsuperscript{(3)}} & \opdbest{21.41}{\scriptsize\color{inkmuted}\,$\pm$8.22\textsuperscript{(3)}} \\
RG-Algorithm & 19.5 & 63.9 & \opdbest{34.71}{\scriptsize\color{inkmuted}\,$\pm$2.84\textsuperscript{(3)}} & 29.11{\scriptsize\color{inkmuted}\,$\pm$3.75\textsuperscript{(3)}} & 31.68{\scriptsize\color{inkmuted}\,$\pm$1.12\textsuperscript{(3)}} \\
\Block[l]{1-6}{\grouplabel{Science}}\rule{0pt}{12.5pt} & & & & & \\[-2.5pt]
GPQA Diamond & 35.9 & 59.1 & 43.13{\scriptsize\color{inkmuted}\,$\pm$1.78\textsuperscript{(10)}} & \opdbest{45.15}{\scriptsize\color{inkmuted}\,$\pm$2.48\textsuperscript{(10)}} & 43.69{\scriptsize\color{inkmuted}\,$\pm$2.27\textsuperscript{(10)}} \\
SuperGPQA & 26.3 & 37.8 & 28.62{\scriptsize\color{inkmuted}\,$\pm$0.86\textsuperscript{(3)}} & 28.56{\scriptsize\color{inkmuted}\,$\pm$1.38\textsuperscript{(3)}} & \opdbest{29.28}{\scriptsize\color{inkmuted}\,$\pm$0.73\textsuperscript{(3)}} \\
SciBench & 26.4 & 57.4 & 35.69{\scriptsize\color{inkmuted}\,$\pm$0.95\textsuperscript{(3)}} & 35.31{\scriptsize\color{inkmuted}\,$\pm$0.46\textsuperscript{(3)}} & \opdbest{35.79}{\scriptsize\color{inkmuted}\,$\pm$0.85\textsuperscript{(3)}} \\
\textbf{Average} & 26.0 & 50.7 & 38.95 & 39.18 & \opdbest{39.30} \\
\CodeAfter
  \begin{tikzpicture}[line cap=butt]
    \draw[rulegray,line width=.4pt] ([xshift=4pt]2-|2) -- ([xshift=-4pt]2-|4);
    \draw[rulegray,line width=.4pt] ([xshift=4pt]2-|4) -- ([xshift=-4pt]2-|6);
    \draw[rulegray,line width=.55pt] (3-|1) -- (3-|7);
    \draw[rulegray,line width=.55pt] (\OPDMainAvgLine-|1) -- (\OPDMainAvgLine-|7);
    \draw[dweoutline,line width=.9pt,rounded corners=5pt]
      ([xshift=-1.2pt,yshift=1.5pt]1-|6) rectangle
      ([xshift=1.2pt,yshift=-2pt]\OPDMainLastLine-|7);
  \end{tikzpicture}
\end{NiceTabular}
\end{table}

\FloatBarrier
\endgroup

\endgroup

\subsubsection{Training Dynamics}
\label{app:opd-training-dynamics}

\begin{figure}[ht]
  \centering
  \setlength{\abovecaptionskip}{4pt}
  \includegraphics[width=\textwidth]{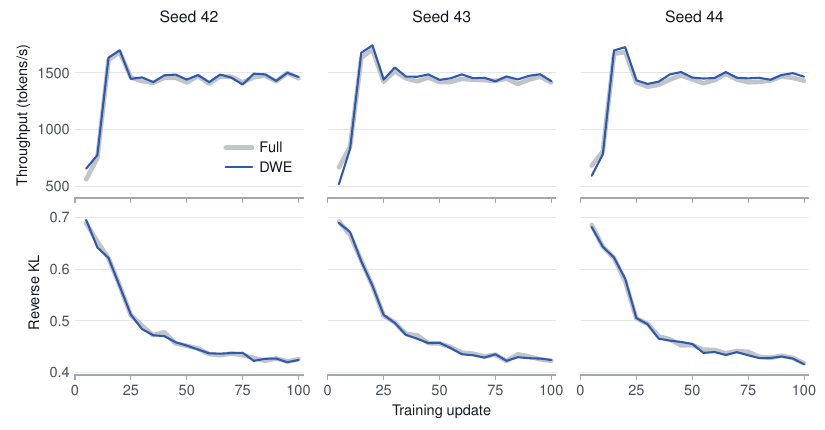}
  \caption{\textbf{Training dynamics.} Full and \method{} across three training seeds.
  Columns correspond to seeds 42, 43, and 44. The top row shows response-token
  throughput; the bottom row shows full-vocabulary reverse KL.
  Curves summarize consecutive blocks of five updates.
  Each method is evaluated on its own responses.}
  \label{fig:opd-training-dynamics}
\end{figure}

Figure~\ref{fig:opd-training-dynamics} shows the throughput and reverse-KL loss
during training for seeds 42, 43, and 44. Each point summarizes five
consecutive updates. Throughput is computed as the total number of response
tokens divided by the total update time, excluding initialization and
checkpoint I/O. Loss is averaged over the five updates.

Across all three seeds, the reverse-KL losses of \method{} and Full decrease
along similar trajectories, while \method{} achieves higher throughput in
most five-update blocks.

\FloatBarrier
\subsubsection{Memory Usage}
\label{app:opd-head-memory}
We measure peak allocated memory during the complete output-head operation
using the fixed inputs described in
Appendix~\ref{app:opd-head-efficiency}. Each method,
domain, and position count runs in a fresh process with three warmup and
three measured repetitions. Table~\ref{tab:opd-head-memory} reports the maximum
increase over pre-forward allocation across repetitions and domains, with
each method's training checkpointing behavior retained.

\method{} reduces incremental peak allocated memory by 4.5--6.0\% across
the three token counts.
For backward computation, \method{} retains at most $2K$ indices (int64) and
coefficients (FP32) per token, totaling
$2K(8+4)=\OPDAdjointBytesPerToken{}$ bytes at $K=8$, or
\OPDAdjointMiBEightK{}\,MiB for 8K positions. Both methods also allocate a dense
output-weight gradient, while \method{} uses vocabulary-wide temporaries
during estimator preparation. The measured peaks account for these
allocations.

\begin{table}[ht]
  \centering
  \begin{minipage}{.88\textwidth}
  \caption{\textbf{Output-head peak memory.} Values report incremental peak
  allocated memory, excluding preallocated model weights and inputs.}
  \label{tab:opd-head-memory}
  \vspace{5pt}
  \centering
  \renewcommand{\arraystretch}{1.25}
  \begin{tabular*}{\linewidth}{@{\extracolsep{\fill}}r r >{\color{resulttext}}r r@{}}
    \toprule
    \multirow{2}{*}{Token positions} & \multicolumn{2}{c}{Peak memory (MiB)} & \multirow{2}{*}{Change} \\
    \cmidrule(lr){2-3}
    & Full & \method{} & \\
    \midrule
1{,}024 & 1942.4 & 1854.2 & $-$4.5\% \\
4{,}096 & 1981.4 & 1878.8 & $-$5.2\% \\
8{,}192 & 2033.5 & 1911.5 & $-$6.0\% \\
    \bottomrule
  \end{tabular*}
  \end{minipage}
\end{table}
\end{document}